\documentclass[letterpaper]{article} %
\usepackage{aaai23}  %
\usepackage{times}  %
\usepackage{helvet}  %
\usepackage{courier}  %
\usepackage[hyphens]{url}  %
\usepackage{graphicx} %
\usepackage{natbib}  %
\usepackage{caption} 
\usepackage{algorithm}
\usepackage{algorithmic}
\usepackage{multirow}
\usepackage{placeins}
\usepackage{newfloat}
\usepackage{listings}
\DeclareCaptionStyle{ruled}{labelfont=normalfont,labelsep=colon,strut=off} %
\floatstyle{ruled}
\newfloat{listing}{tb}{lst}{}
\floatname{listing}{Listing}
\usepackage{multibib}
\newcites{A}{Appendix References}

\title{Weather2vec: Representation Learning for Causal Inference with Non-Local Confounding in Air Pollution and Climate Studies}
\author{
    Mauricio Tec\textsuperscript{\rm 1}\\
    James G. Scott\textsuperscript{\rm 1}\textsuperscript{\rm 2},
    Corwin M. Zigler\textsuperscript{\rm 3}
}
\affiliations{
    \textsuperscript{\rm 1}Department of Biostatistics, Harvard T.H. Chan School of Public Health\\
    \textsuperscript{\rm 2}Department of Statistics and Data Sciences,     The University of Texas at Austin\\
    \textsuperscript{\rm 3}Department of Information, Risk, and Operations Management, The University of Texas at Austin \\
    mauriciogtec@hsph.harvard.edu, james.scott@mccombs.utexas.edu, cory.zigler@austin.utexas.edu
}

\usepackage{amsmath,amsthm,amsfonts,bm}

\newcommand{\indep}{\perp \!\!\!\! \perp}

\def\eqref#1{equation~\ref{#1}}

\def\1{\bm{1}}

\def\va{{\bm{a}}}

\def\vgamma{{\bm{\gamma}}}

\def\mL{{\bm{L}}}

\def\mX{{\bm{X}}}
\def\mY{{\bm{Y}}}
\def\mZ{{\bm{Z}}}

\def\mGamma{{\bm{\Gamma}}}

\DeclareMathAlphabet{\mathsfit}{\encodingdefault}{\sfdefault}{m}{sl}
\SetMathAlphabet{\mathsfit}{bold}{\encodingdefault}{\sfdefault}{bx}{n}
\newcommand{\tens}[1]{\bm{\mathsfit{#1}}}
\def\tA{{\tens{A}}}

\def\tF{{\tens{F}}}

\def\tX{{\tens{X}}}

\def\tZ{{\tens{Z}}}

\def\gL{{\mathcal{L}}}

\def\gN{{\mathcal{N}}}

\def\gR{{\mathcal{R}}}

\def\sG{{\mathbb{G}}}

\def\sS{{\mathbb{S}}}

\newcommand{\E}{\mathbb{E}}

\newcommand{\R}{\mathbb{R}}

\newtheorem{assumption}{Assumption}
\newtheorem{proposition}{Proposition}
\newtheorem{mydef}{Definition}

\usepackage{xspace}
\usepackage{enumerate,enumitem}
\usepackage{subcaption}

\usepackage[capitalize]{cleveref}

\newcommand{\method}{{\em weather2vec}\xspace}
\newcommand{\SO}{\textrm{SO}}

\newcommand{\covar}{\mX_s}
\newcommand{\covardelta}{\mX_{s + \delta}}
\newcommand{\covarprime}{\mX_{s'}}
\newcommand{\propconf}{\mL_s}
\newcommand{\conf}{\mL_s}
\newcommand{\lat}{\mZ_{\theta, s}}
\newcommand{\out}{Y_s}
\newcommand{\treat}{A_s}
\newcommand{\treatprime}{A_{s'}}
\newcommand{\neigh}{\gN_s}
\newcommand{\neighminus}{\gN_s\setminus \{s\}}
\newcommand{\Xneigh}{\tX_{\gN_s}}
\newcommand{\Aneigh}{\tA_{\gN_s}}
\newcommand{\aneigh}{\va_{\gN_s}}

\newcommand{\Ind}{\mathbb{I}}

\begin{document}

\maketitle

\begin{abstract}
Estimating the causal effects of a spatially-varying intervention on a spatially-varying outcome may be subject to non-local confounding (NLC), a phenomenon that can bias estimates when the treatments and outcomes of a given unit are dictated in part by the covariates of other nearby units. In particular, NLC is a challenge for evaluating the effects of environmental policies and climate events on health-related outcomes such as air pollution exposure. This paper first formalizes NLC using the potential outcomes framework, providing a comparison with the related phenomenon of causal interference. Then, it proposes a broadly applicable framework, termed \emph{weather2vec}, that uses the theory of balancing scores to learn representations of non-local information into a scalar or vector defined for each observational unit, which is subsequently used to adjust for confounding in conjunction with causal inference methods. The framework is evaluated in a simulation study and two case studies on air pollution where the weather is an (inherently regional) known confounder.
\end{abstract}

\section{Introduction}\label{sec:intro}

Causal effects of spatially-varying exposures on spatially-varying outcomes may be subject to \emph{non-local confounding} (NLC), which occurs when the treatments and outcomes for a given unit are affected by \emph{covariates} of other nearby units \citep{cohen2008obesity, florax1992specification, chaix2010neighborhood, elhorst2010applied}. In simple cases, NLC can be resolved using simple summaries of non-local data, such as the averages of the covariates over pre-specified neighborhoods. But in many realistic settings, NLC is caused by the complex interaction of spatial factors, and thus it cannot be resolved using simple \emph{ad hoc} summaries of neighboring covariates. For such scenarios, we propose \method, a framework that uses a U-net \citep{ronneberger2015u} to learn representations that encode NLC information and can be used in conjunction with standard causal inference tools\footnote{The code for reproducibility is available at \url{https://github.com/mauriciogtec/weather2vec-reproduce}}. The method is broadly applicable to settings where the covariates are available over a grid of spatial units, and where the outcome and treatment are observed in some subset of the grid. 

The name \method stems from its motivation to address limitations in current methods for estimating causal effects in environmental studies where meteorological processes are known confounders, aiming to contribute to the development of new flexible machine learning tools to assess the \emph{causal effect} of policies and climate-related events on health-relevant outcomes: a task which has been recently identified by \citet{rolnick2022tackling} as a pressing outstanding challenge for tackling the effects of climate change.

Two applications will be discussed in detail. The first application follows an earlier analysis by \citet{papadogeorgou2019adjusting}, who estimated the air quality impact of power plant emissions controls. This case study evaluates the method's ability to reduce NLC under sparsely observed treatments (in combination with with propensity matching methods \citep{rubin2005causal}).
The second application is in \emph{meteorological detrending} \cite{wells2021improved}, and uses \method to deconvolve climate variability from policy changes when characterizing long-term air quality trends.
These two examples are accompanied by a simulation study comparing alternative adjustments to account for NLC. 

In summary, this article has three aims:
\begin{enumerate}[itemsep=0pt, parsep=0pt]
  \item Provide a rigorous characterization of NLC using the potential outcomes framework, clarifying some connections with causal interference.
  \item Expand the library of NN methods in causal inference by proposing a U-net as a viable model to account for NLC in conjunction with standard causal inference tools.
   \item Establish a promising research direction for addressing NLC in scientific studies of air pollution exposure -- in which NLC is a common problem (driven by meteorology) for which widely applicable tools are lacking. 
\end{enumerate}

We investigate two mechanisms to obtain the representations: one supervised, and one self-supervised.  The supervised one formally links the representation of NLC to the balancing property of propensity (and prognostic) scores in the causal inference literature \citep{rubin_for_2008,hansen2008prognostic}. This approach requires that the outcome and treatment are densely available throughout the covariates' grid. By contrast, the self-supervised approach first learns representations encoding neighboring covariate information into a low-dimensional vector, which can subsequently be included as confounders in downstream causal analyses when the outcomes and treatments are sparsely observed on the grid.

\paragraph{Related work} 
Previous research has investigated NNs for the (non-spatial) estimation of balancing scores \citep{keller2015neural, westreich2010propensity, setoguchi2008evaluating} and counterfactual estimation  \citep{shalit2017estimating, johansson2016learning, shi2019adapting}.
But none of these works specifically consider NLC. 

Relevant applications of U-nets in environmental studies include forecasting \citep{larraondo2019data, sadeghi2020improving}, estimating spatial data distributions from satellite images \citep{hanna2021multitask,fan2021resolving}, indicating that U-nets are powerful tools to manipulate rasterized weather data. Also relevant, \citet{lu2005meteorologically} give a specific application of NNs for meteorological detrending, although without considering adjusting for neighboring covariates.
\citet{shen2017influence} do consider regional dependencies by applying patch-wise PCA to extract meteorological features to improve prediction of air pollution. Approaches to learn summaries of neighboring covariates for regression-based causal inference have been investigated in the econometrics literature. For example, WX-regression models \citep{elhorst2010applied} formulate the outcome as a linear function of the treatment and the covariates of some pre-specified neighborhood.  CRAE \citep{blier2020encoding} uses an autoencoder over pre-extracted patches of regional census data that is fed into an econometric regression. In contrast to predictive regression-based approaches, \method aims at learning balancing scores, which have known benefits that include the ability to empirically assess the threat of residual confounding and offer protection against model misspecification that arises when modeling outcomes directly \citep{rubin_for_2008}. 

There is also a maturing literature on adjusting for unobserved spatially-varying confounding \citep{reich2021review}.
Spatial random effect methods are popular in practice, although \citet{khan2020restricted} have highlighted their sensitivity to misspecification for the purposes of confounding adjustment. The distance adjusted propensity score matching (DAPSm) \citep{papadogeorgou2019adjusting} matches units based jointly on estimated propensity scores and spatial proximity under the rationale that spatial proximity can serve as a proxy for similarity in spatially-varying covariates. In the same spirit,
\citet{veitch2019using} use graph embeddings to account for proximity within a network as a proxy for confounding.
In general, the primary target of spatial confounding methods are settings where confounding is local conditional on the unobserved spatially-varying confounders--in contrast to NLC.

Finally, NLC is distinct from \emph{causal interference} \citep{tchetgen2012causal, forastiere2021identification, sobel2006randomized, zigler2021bipartite, ogburn2014causal,bhattacharya2020causal}, although both phenomena arise from spatial (or network) interaction, and they both impose limitations on standard causal inference methods. While forms of NLC have often been acknowledge in the literature of interference, to the best of our knowledge, flexible statistical methods specifically addressing NLC by learning the dependencies with respect to neighboring covariates do not exist.

\section{Potential outcomes and NLC}\label{sec:rubin}

We now recall the potential outcomes framework, also known as the Rubin Causal Model (RCM) \citep{rubin_for_2008}, and we later adapt it to the case of NLC confounding. The RCM distinguishes between the observed outcome $\out$ at unit $s$ and those that would be observed under counterfactual (potential) treatments $Y_s(a)$ (formally defined below). We start with some notation.  The assigned treatment is denoted $\treat$. It is assumed to be binary for ease of presentation, although the ideas generalize to more general treatments. For instance, in our Application 1 the treatment is whether or not a catalytic device is installed on a power plant to reduce the emissions of some pollutant. $\sS$ is the set where the outcome and treatment are measured (e.g., the location of the power plants); $\sG \supset \sS$ is a grid containing the rasterized covariates $\{\covar \in \mathbb{R}^d \colon s\in\sG\}$; for any $B\subset\sG$, $\tX_{B}= \{\covar \mid s\in B\}$; $X \indep Y \mid Z$ denotes conditional independence of $X$ and $Y$ given $Z$; lastly,  $p(\cdot)$ denotes a generic probability or density function. We will assume throughout that $\covar$ only contains pre-treatment covariates, meaning they are not affected by the treatment or outcome.

\begin{mydef}[Potential outcomes]\label{def:potential-outcomes} The potential outcome $\out(\va)$ is the outcome value that \textit{would} be observed at location $s$ under the global treatment assignment $\va=(a_1,\hdots, a_{|\sS|})$.
\end{mydef}

For $\out(\va)$ to depend only on $a_s$, the RCM needs an additional condition called the \textit{stable unit treatment value assumption}, widely known as SUTVA, and encompassing notions of {\it consistency} and ruling out {\it interference}.

\begin{assumption}[SUTVA] \label{def:sutva} (1) Consistency: there is only one version of the treatment. (2) No interference: the potential outcomes for one location do not depend on treatments of other locations. Together, these conditions imply that $\out(\va)=\out(a_s)$ for any assignment vector $\va\in\{0,1\}^{|\sS|}$, and that the observed outcome is the potential outcome for the observed treatment, i.e., $\out=\out(\treat)$.
\end{assumption}

To contextualize SUTVA in our power plant example, observe that it would be violated if the pollution measured at $s$ depends not only on whether or not the catalytic device was installed at that power plant (that is, on the assignment $A_s$), but also on whether or not the device was installed on other power plants ($A_{s'}$ for $s\neq s$). We assume SUTVA throughout as it is common in many causal inference studies. Then, the potential outcomes allow to define an important estimand of interest: the average treatment effect.

\begin{mydef}[ATE]\label{def:ate} The average treatment effect (ATE) is the quantity
$\tau_\text{ATE} = |\sS|^{-1}\textstyle{\sum_{s\in\sS}} \left\{\out(1) - \out(0)\right\}$.
\end{mydef}

One cannot estimate the ATE directly since one never simultaneously observes $\out(0)$ and $\out(1)$. The next assumption in the RCM formalizes conditions for estimating the ATE, (or other causal estimands) with observed data by stating that any  observed association between $\treat$ and $\out$ is not due to an unobserved factor.

\begin{assumption}[Treatment Ignorability]\label{def:ignorability} The treatment $\treat$ is ignorable with respect to some vector of controls $\propconf$ if and only if $\out(1), \out(0) \indep \treat \mid \propconf$.
\end{assumption}

\begin{figure}[htb]
    \centering
    \begin{subfigure}[t]{.3\linewidth}\centering
    \includegraphics[width=.85\linewidth]{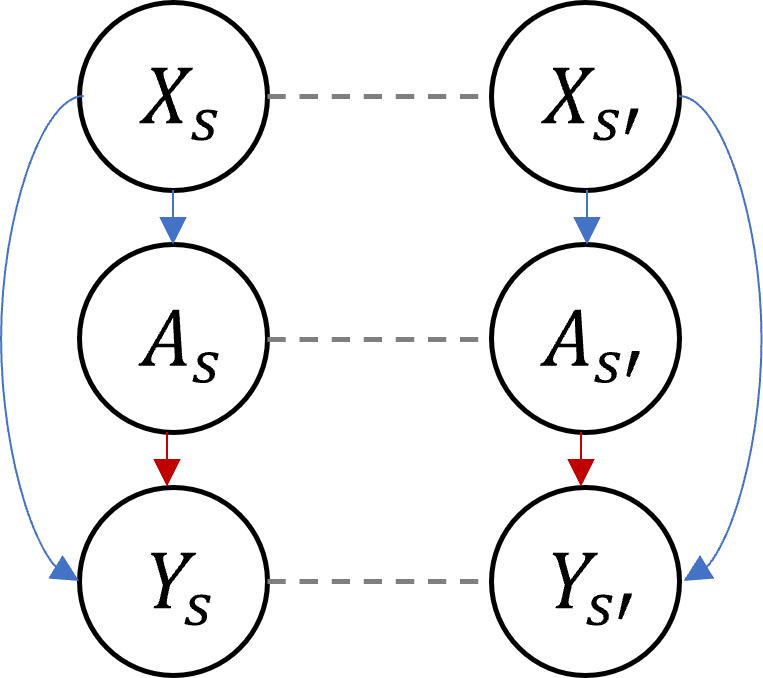}
    \subcaption{ Local Confounding.}
    \label{fig:confounding-types:local}
    \end{subfigure}
    \begin{subfigure}[t]{.3\linewidth}\centering
    \includegraphics[width=.85\linewidth]{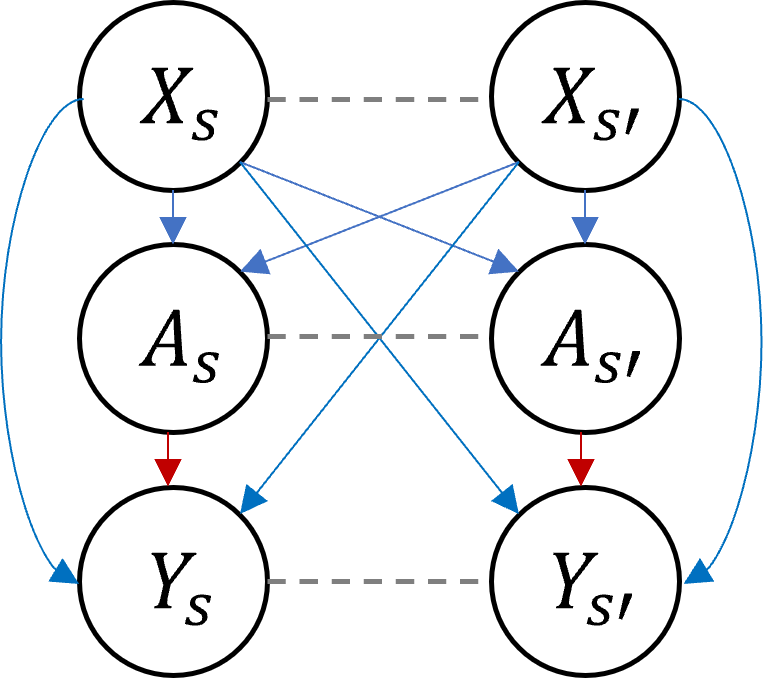}
    \subcaption{ NLC, no interference.}
    \label{fig:confounding-types:nlc}
    \end{subfigure}
    \begin{subfigure}[t]{.3\linewidth}\centering
    \includegraphics[width=.85\linewidth]{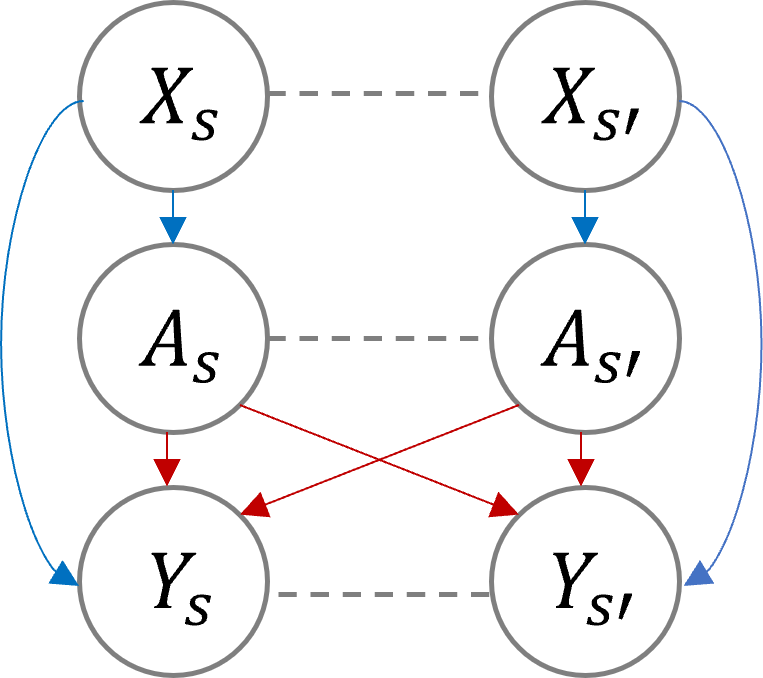}
    \subcaption{ Interference, no NLC.}
    \label{fig:confounding-types:interf}
    \end{subfigure}
    \caption{Confounding types.}
    \label{fig:confounding-types}
    \vskip-12pt
\end{figure}

\vskip-12pt
This ignorability assumption would fails where there exist unobserved confounders. For the sake of brevity, we will say that $\propconf$ is \emph{sufficient} to mean that the treatment is ignorable conditional on $\propconf$. We now introduce NLC, which occurs when non-local covariates are among the confounders. It is formally stated as follows:

\begin{mydef}[Non-local confounding] We say there is non-local confounding (NLC) when there exist neighborhoods $\{\neigh \subset \sG \mid s\in\sS \}$ such that $\propconf=\Xneigh$ is sufficient and the neighborhoods are necessarily non-trivial ($\neigh\neq\{s\}$).
\label{def:nlc}\end{mydef}

In our power plant example, atmospheric vectors $X_{s'}$ are associated with the air pollution outcomes at other locations $Y_{s}$ \citep{shen2017influence}, as well as the probability of installing a catalytic device on a power plant, $A_{s}$.
\cref{fig:confounding-types:local,fig:confounding-types:nlc} show a graphical representation of local confounding versus NLC.  Horizontal dotted lines emphasize spatial correlations in the covariate, treatment, and outcome processes that do not result in confounding. For contrast, \cref{fig:confounding-types:interf} shows the distinct phenomenon of (direct) interference, in which $A_{s'}$ affects $A_{s}$ \citep{ogburn2014causal}. (This depiction of is only one of the forms that interference can take. For instance, it may also happen through contagion \citep{ogburn2014causal}.) 
Interference is further discussed in \cref{sec:interference}.

Subsequent discussion of the size of the NLC neighborhood, $\neigh$, will make use of the following proposition stating that a neighborhood containing sufficient confounders can be enlarged without sacrificing the sufficiency.

\begin{proposition}\label{prop:pre-treatment}
Let $\propconf$ be a sufficient set of controls including only pre-treatment covariates. and let $\propconf'$ be another set of controls satisfying $\propconf' \supset \propconf$. Then, $\propconf'$ is also sufficient.
\end{proposition}

All the proofs are in Appendix \ref{appendix:proofs}. We can now state a classic result regarding the identifiability of causal effects from observed data under the above assumptions.

\begin{proposition}\label{prop:unbiasedness} Assume SUTVA holds and that $\propconf$ is sufficient. Then
\begin{equation}\label{eq:ate}
  \begin{aligned}
      \E\left[\E[\out \mid \propconf, \treat=1] - \E[\out \mid \propconf, \treat=0]\right],
  \end{aligned}
\end{equation}
is an unbiased estimator of $ \tau_\text{ATE}$ (where $s$ is taken uniformly at random from $\sS$).
\end{proposition}
\cref{eq:ate} already offers a way to estimate causal effects from observed data (by estimating the two inner conditional expectations). However, it can be highly sensitive to the specification of the expected outcome model. There are many alternative, one which is to use inverse probability of treatment weighting (IPTW) \citep{cole2008constructing}; described in \cref{app:iptw} for completeness, and which an estimate of the \emph{propensity score}, introduced in the next section.

\section{Adjustment for NLC with \method}

Accounting for NLC would be fairly straightforward provided infinite data and the right set of confounders.  By virtue of Proposition \ref{prop:pre-treatment}, one could, in principle, specify a non-linear regression $Y_s \approx f(A_s, \tX_\sG, s)$ that includes every non-local covariate $
\covarprime\in \tX_{\sG}$ as part of the regressors. With large model capacity, infinite repeated samples per location, this regression would perfectly estimate $\E[\out \mid \propconf, \treat=a]$, and thus be able to estimate the ATE using Proposition \ref{prop:unbiasedness}. But this scenario is far from realistic. Most commonly, there will be only one observation for each $s$, requiring additional structure to enable statistical estimation.
Thus, we consider the question: what kind of statistical and functional model (e.g., to predict the probability of treatment) reflects the causal structure of NLC and allows for flexible statistical models under such restrictions?

One desirable statistical property to consider is \emph{spatial stationarity}. Intuitively, it entails that the distributions of $\out$ and $\treat$ with respect to a neighboring covariate $\covarprime$ should only depend on $\delta=s-s'$ (their relative position). Formally, it requires that for any set $B\subset \sG$, displacement vector $\delta$, and $s\in \sG$, the following identity holds $p(\treat, \out \mid \tX_B=x)\overset{}{=} p(A_{s + \delta}, Y_{s + \delta} \mid \tX_{B + \delta}=x)$. For \method, we focus on the the U-net \citep{ronneberger2015u}, which are neural network models composed of convolution operators they are approximately spatially stationary and allow to learn predictions from neighboring inputs at every point of a grid. An overview of U-nets is provided in the next section for completeness. A key property is that a U-net $f_\theta$ can transform the input covariates $\tX_\sG$ onto an output grid $\tZ_{\theta,\sG}:=f_\theta(\tX_\sG)$ of same spatial dimensions in which each scalar or vector $\lat\in \tZ_{\theta,\sG}$ localizes contextual spatial information from the input grid.  

U-nets are not the only neural architecture with these properties. For instance, one could adapt residual networks \citep{he2016deep} as a shallow alternative to a U-net.
The essence of \method is to define appropriate learning tasks to obtain the NN weights $\theta$. Two such tasks are considered, summarized below and in \cref{sec:algorithms}, and described in detail in subsequent sections.

\begin{enumerate}[itemsep=0pt, itemindent=0pt, parsep=0pt]
  \item (\textbf{Supervised)}\quad Assuming the treatment is densely available over $\sG$, estimate $\tZ_{\theta,\sG}$ as the probability of treatment conditional on non-local covariates.
  \item (\textbf{Self-supervised)}\quad If the treatment is not densely available over $\sG$, then learn $\lat$ so that it is highly predictive of $\covarprime$ for any $s'$ within a specified radius of $s$. Then use $\lat$ as an input in a second-stage model to learn the treatment probability.
\end{enumerate}

These strategies allow learning a \emph{propensity score}, $p(\treat=1 \mid \lat)$, which can be used within a well-established causal inference technique such as IPTW (Appendix \ref{app:iptw}) to produce robust causal estimates of $\tau_\text{ATE}$. The key innovation with respect to traditional causal propensity score methods is the inclusion of NLC information. Later we also consider a variant based on \emph{prognostic scores} \citep{hansen2008prognostic}, which are predictive functions of the untreated outcomes.

\subsection{An overview of the U-net for summarizing NLC}

The U-net transformation involves two parts: a \emph{contractive} stage and a symmetric \emph{expansive stage}. These steps use convolutions with learnable parameters and non-linear functions to aggregate information from the input grid spatially and create rich high-level features. The convolutions in the contractive path duplicate the number of latent features at each layer. Then, these intermediate outputs go through \textit{pooling} layers which halve the spatial dimensions. Together, these operations augment the dimensionality of each point of the grid, combining information at many spatial points with richer information contained at fewer points. Convolutions propagate information spatially, and the deeper they are in the contractive path, the larger their propagation reach (in the original scale of the input grid).
The expansive path, on the other hand, uses \textit{up-sampling} to progressively interpolate the deep higher-level features back to a finer spatial lattice, and then uses convolutions to reduce back the latent dimensionality at each grid point; with the characteristic that, in contrast to the input grid, every point now localizes spatial information. The output vector can have any arbitrary dimension after possibly applying an additional layer after the expansive path (or before the contractive path, or both). \cref{fig:architecture} in the Appendix provides a visual example of the U-net architecture. See the original work by \citet{ronneberger2015u} for additional details on the U-net. 

The unknown weights $\theta$ learn what non-local information is summarized by $\lat$. The depth of the U-net (number of down/up layers) dictates the maximum radius of spatial aggregation. Shallow U-nets operating on fine-grained grids may have limited spatial aggregation capabilities. 
Convolutions, pointwise activations, pooling, and upsampling layers are all spatially stationary operations. However, some commonly used operations, such as padding and batch normalization layers, may affect stationarity. Some strategies that can be implemented to reduce their impact is removing padding, masking outputs, and replacing batch normalization with FRN layers \citep{singh2020filter} or other valid normalization. We implement these strategies further in the details of our applications.

\subsection{Learning NLC representations via supervision}\label{sec:supervised}

The supervised approach links the proposed representation learning to the seminal work of \citet{rubin1978bayesian} on propensity scores for causal inference. We briefly summarize balancing scores following the standard presentation. 

  \begin{mydef}[Propensity score]\label{def:balancing} $b(\propconf)$ is a {\em balancing score} iff $\treat \indep \propconf \mid b(\propconf)$. The coarsest balancing score is $b(\propconf) := p(\treat=1 \mid \propconf)$, widely known as the \emph{propensity score}.
  \end{mydef}
  
  \begin{mydef}[Prognostic score]\label{def:prognostic}  $b(\propconf)$ is a {\em prognostic score} iff $\out(0) \indep \mL_s \mid b(\propconf)$. The coarsest prognostic score is $b(\propconf) := \E[\out(0) \mid \propconf]$. 
  \end{mydef}

  The propensity score blocks confounding through the treatment \citep{rubin2005causal}; prognostic scores do so through the outcome \citep{hansen2008prognostic}. The importance of these definitions is summarized by the next well-known result.
  
  \begin{proposition}\label{prop:bs-reduction}
  If $b(\propconf)$ is a balancing score, then $\propconf$ suffices to control for confounding iff $b(\propconf)$ does. The same result holds for the prognostic score under the additional assumption of no effect modification.
  \end{proposition}

This result suggests to consider $\propconf$ to be implicitly defined by the full covariates grid ``centered" at $s$, letting the network weights learn the effective radius of dependence. We can equate $\mZ_{\theta,s}$ to either the propensity score or the prognostic score via direct regression, which amounts to minimizing the binary classificaiton and regression loss:
\begin{align}
\gL_\text{sup}^\text{prop}(\theta)&= 
\textstyle{\sum_{s\in \sS}} \text{CrossEnt}(A_s, \mZ_{\theta,s}) \label{eq:sup-prop} \\
\gL_\text{sup}^\text{prog}(\theta) &=
\textstyle{\sum_{s\in \sS: \treat = 0}} (\out - \mZ_{\theta,s})^2 \label{eq:sup-prog}.
\end{align}
Notice that \cref{eq:sup-prog} applies only to untreated units. The learned propensity score can be directly plugged into a robust estimator such as IPTW (\cref{app:iptw}) or it can be used as a covariate in the case of the prognostic score. Learning $\theta$ through supervision results in an efficient scalar $\lat$ compressing NLC information, allowing for $\theta$ to just attend to relevant neighboring covariate information that pertains to confounding.  Yet supervision may not be possible with small-data studies where $\out$ and $\treat$ are only measured sparsely. In such cases, the supervised model will likely overfit the data. For example, in application 1 in section \ref{sec:applications}, $\sS$ consists only of measurements at 473 power plants, while the size of $\sG$ is $128 \times 256$. Overfitting would invalidate common causal inference methods like IPTW that rely on unbiased estimates of the propensity score.

\subsection{Representations via self-supervised dimensionality reduction}\label{sec:self-supervised}

Self-supervision frames the representation learning problem as dimension reduction without reference to the treatment or outcome. The representations are then used to learn a balancing score for causal effect estimation in a second analysis stage.  This approach requires specification of a fixed neighborhood $\gN_s$ (parameterized by a radius $R$) and latent dimension $k$, resulting on different representations for different hyper-parameter choices, which can be selected using standard model selection techniques (such as AIC) in the second stage. The dimension reduction's objective is that $\lat$ encodes predictive information of any $\covardelta$ for $(s + \delta) \in \neigh$.
A simple predictive model $\covardelta \approx g_\phi(\lat, \delta)$ is proposed. First, let $\mGamma_{\phi}(\cdot)$ be a function taking an offset $\delta$ as an input and yielding a $k\times k$ matrix, and let $h_\psi(\cdot)\colon \R^k \to \R^d$ be a decoder with output values in the covariate space. The idea is to consider $\mGamma_{\phi}(\delta)$ as a selection operator acting on $\lat$. The task loss function can be written succinctly as
\begin{equation}\label{eq:self}
\begin{aligned}
\gL_\text{self}(\theta, \phi,\psi\mid R) &= \\
\textstyle{\sum_{s\in \sG}\sum_{ \{\delta\colon \lVert \delta \rVert \leq R\}}} & \left(\covardelta - h_\psi(  \mGamma_\phi(\delta)\lat)\right)^2.
\end{aligned}
\end{equation}
\cref{appendix:motivation-selfsup} provides additional intuition about \cref{eq:self}. A connection with PCA is also described in \cref{appendix:pca}. While \cref{eq:self} is formulated for spatial dimensionality reduction only, an advantage of this expression is that it can be easily extended to multi-task settings and dimensionality reduction in the temporal axis for spatiotemporal data. We plan to explore these possibilities for future work.

\section{NLC and Interference}\label{sec:interference}

Section \ref{sec:intro} briefly contrasted NLC with the related problem of interference, a topic that we expand here. We first formalize the concept of interference, following closely the form of interference considered in \citet{forastiere2021identification}, which replaces SUTVA with the following neighborhood-level assumption, termed the \textit{stable unit neighborhood treatment value assignment} (SUTNVA).

\begin{figure}
    \centering
    \includegraphics[width=.26\linewidth]{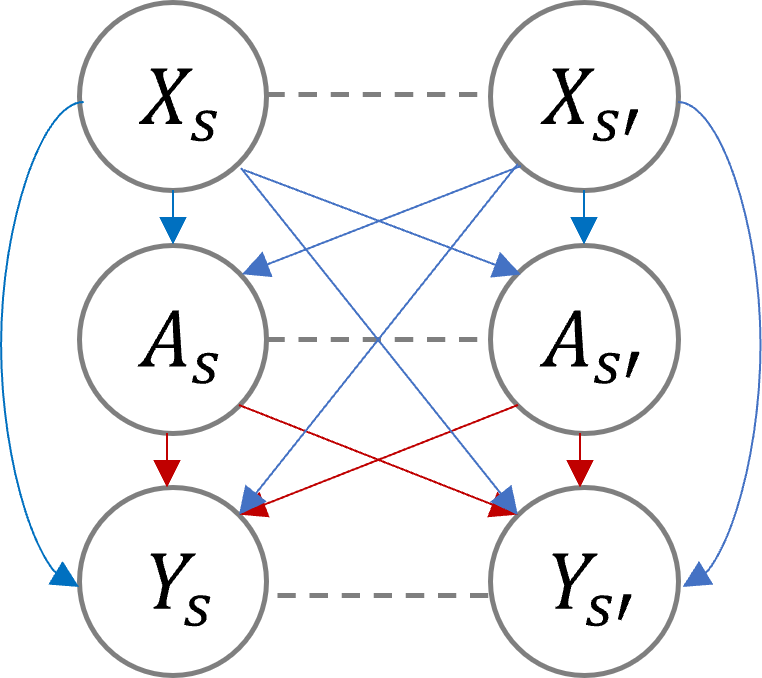}
    \caption{ Interference + NLC.}
    \vspace*{-12pt}
    \label{fig:interf2}
\end{figure}

\begin{assumption}[SUTNVA] \label{def:sutnva} (1) Consistency: there is only one version the treatment. (2) Neighborhood-level interference: for each location $s$, there is a neighborhood $\neigh$ such that the potential outcomes depend only on the treatments at $\neigh$. Together, these conditions imply that $\out(\va)=\out(\aneigh)$ for any assignment vector $\va\in\{0,1\}^{|\sS|}$, and that the observed outcome is the potential outcome for the observed treatment, i.e., $\out=\out(\Aneigh)$.
\end{assumption}
This definition of interference only considers \emph{direct} interference, leaving aside indirect mechanisms such as contagion \citep{ogburn2018challenges, shalizi2011homophily}. Investigating the role of NLC in such scenarios is left for future work. We now describe one generalization of the ATE for this type of direct interference.  The statement uses potential outcomes of the form  $\out(a_s=a, \tA_{\neighminus})$ -- a short-hand notation for the potential outcome that assigns the treatments of all the neighbors of $s$ to their observed treatments in the data.
\begin{mydef}[DATE] The direct average treatment effect (DATE) is the quantity $
\tau_\text{DATE} =  |\sS|^{-1}\textstyle{\sum_{s\in\sS}} \{\out(a_s=1, \tA_{\neighminus})  -\out(a_s=0, \tA_{\neighminus}) \}.$
\end{mydef}
The following proposition  by \citet{forastiere2021identification} states two conditions under which one can ``ignore" interference.
\begin{proposition}\label{prop:interference} Assume SUTNVA. Conditions (1) and (2) correspond to the notions of neighborhood-level ignorability and conditional independence of the neighboring treatments.
If (1) $\Aneigh \indep \out(\va) \mid \conf$  for all $\va\in\{0,1\}^{|\neigh|}$ and (2) $\treat \indep \treatprime \mid \conf$ for all $s\in\sS, s'\in \neigh$. Then
\cref{eq:ate} is an unbiased estimator of $\tau_\text{DATE}.$
\end{proposition}
When NLC is present (the arrows from $X_{s'}$ in \cref{fig:confounding-types:nlc}), conditions (1) and (2) can be violated. To see this, consider \cref{fig:interf2} representing the co-occurrence of interference and NLC. Adjusting only for local covariates would violate condition (1) with a spurious correlation between $Y_s$ and $A_{s'}$ (through the backdoor path $\out \leftarrow \covarprime \to \treatprime$). Similarly, a spurious correlation between $\treat$ and $\treatprime$ would persist via the the path $\treat \leftarrow \covarprime \to \treatprime$. For such cases, \method can play an important role in satisfying (1) and (2) since, after controlling for NLC (consisting in \cref{fig:interf2} of adjusting for both $\covar$ and $\covarprime$ and blocking the incoming arrows from neighboring covariates into one's treatments and outcomes), the residual dependencies would more closely resemble those of \cref{fig:confounding-types:interf}. In summary, adjusting for NLC with \method can aid satisfaction of the conditional independencies required to estimate causal effects with the same estimator used to estimate the ATE absent interference. Notice that $\tau_\text{DATE}$ is not the only estimand of interest, for instance, in future work we wish to explore the role of non-local covariates when estimating \emph{spill-over} effects \citep{ogburn2018challenges}.

\begin{table*}[htb]
\centering
\begin{tabular}{c|c|c|c|c|c|c|c|c|c}
 & \multicolumn{2}{c|}{\emph{patch-based}} & \emph{self-supervised}  &\multicolumn{4}{c|}{\emph{supervised}} & \multicolumn{2}{c}{\emph{spatial + supervised}} 
 \\
\emph{Task} & \textsc{vae}  & \textsc{crae} & \textsc{w2v-self} & 
\textsc{wx} & \textsc{local} & \textsc{avs} &
\textsc{w2v-sup} & \textsc{car} & \textsc{w2v-car}
\\  \hline
 \emph{Linear}  & %
 0.58 & 0.04 & 0.02 &  0.58 & 0.58 & 0.53 & \textbf{0.01} & 0.58 & 0.04  \\
\emph{Linear-sparse}  &
 0.59 & \textbf{0.06} & \textbf{0.06} & 0.58 & 0.58 & 0.57 & 0.1 & 0.59 & 0.1 \\
\hline
\emph{Non-linear} & %
 0.58 & 0.13 & 0.11 & 0.58 & 0.58 & 0.58 & \textbf{0.07} & 0.58 & 0.18  \\
\emph{Non-linear-sparse} &
0.59 & 0.15 & \textbf{0.14} & 0.57 & 0.57 & 0.57 & 0.15 & 0.58 & 0.19 \\
\end{tabular}%
\caption{Comparisons in average causal effect error ($\textit{Bias}=\sum_{i=1}^n n^{-1}(\hat{\tau}_\text{IPTW}^{(i)}-\tau_\text{ATE})|$) for different propensity score models in simulated datasets across $n=10$ random seeds. \emph{Dense task}: $\treat$ and $\out$ are observed on the full $128\times 256$ grid. \emph{Sparse task}: $\treat$ and $\out$ are observed in 1000 points scattered throughout the grid.
} 
\label{tab:simulation}
\end{table*}

\begin{figure*}[htb]
  \centering
  \begin{subfigure}[t]{.31\textwidth}
    \centering
    \includegraphics[width=0.99\textwidth]{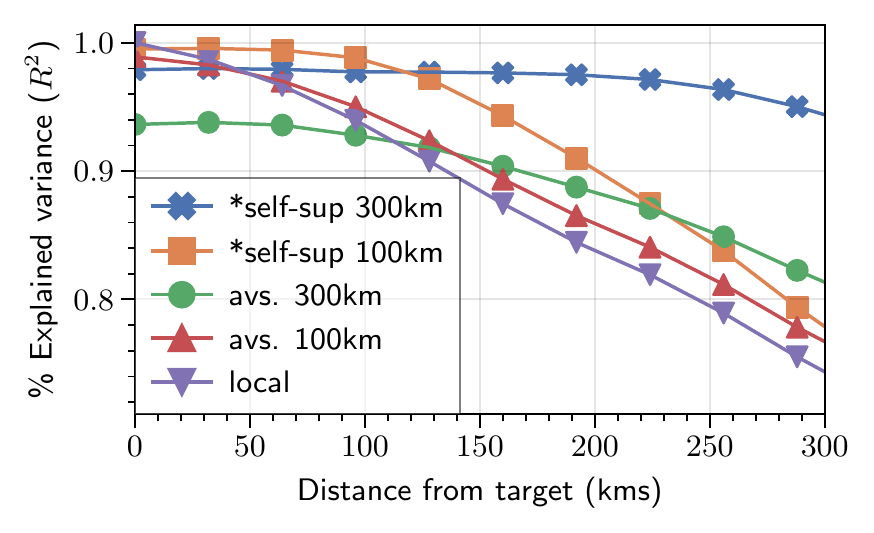}
   \caption{Self-supervision, NARR data}
   \label{fig:app1-self-supervision} 
  \end{subfigure}%
  \begin{subfigure}[t]{.36\textwidth}
    \centering
    \includegraphics[width=0.99\textwidth]{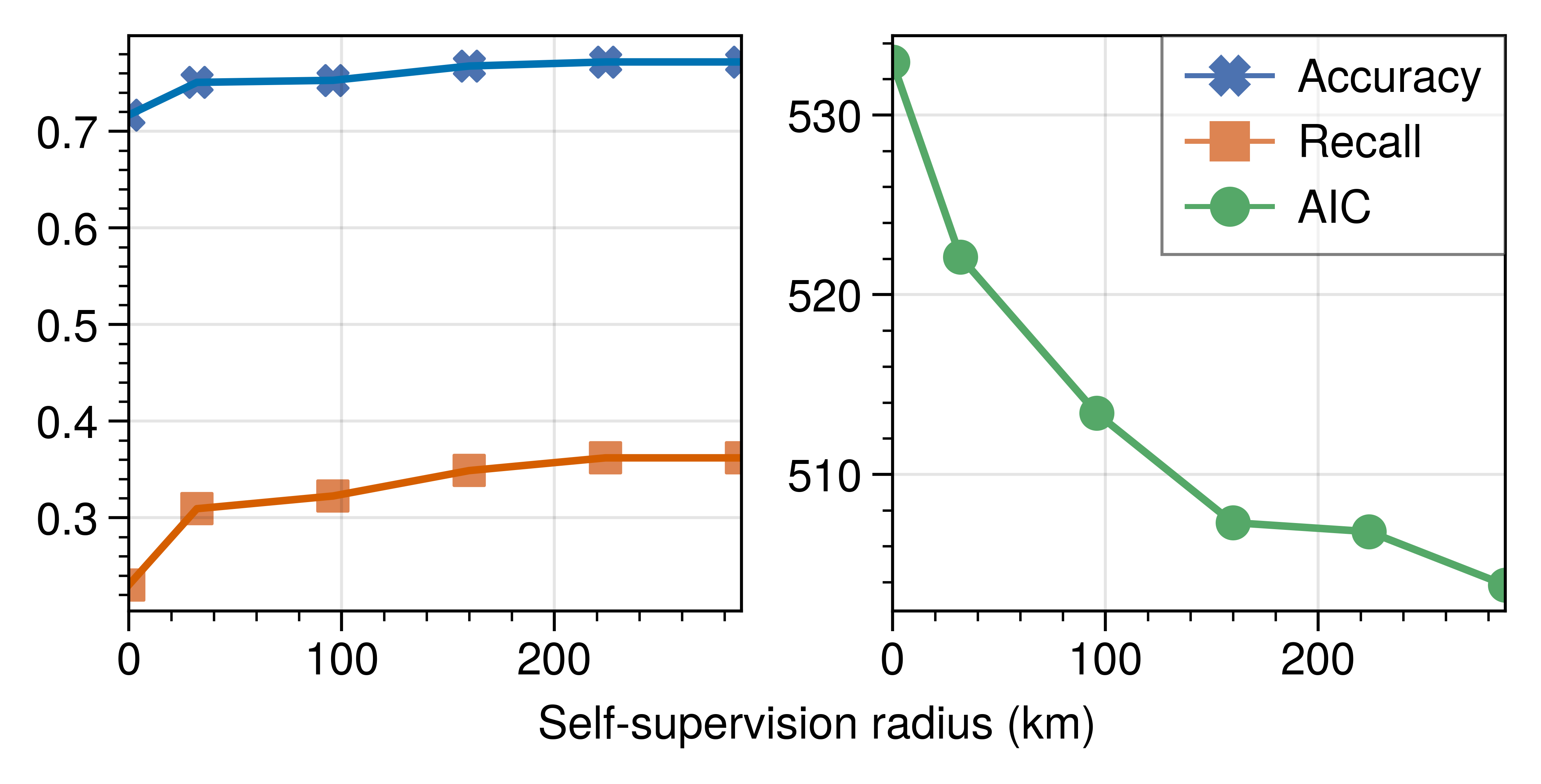}
    \caption{Fit metrics, propensity score model}
    \label{fig:app1-fit-metrics}
  \end{subfigure}%
  \begin{subfigure}[t]{.27\textwidth} \centering
    \includegraphics[width=0.99\textwidth]{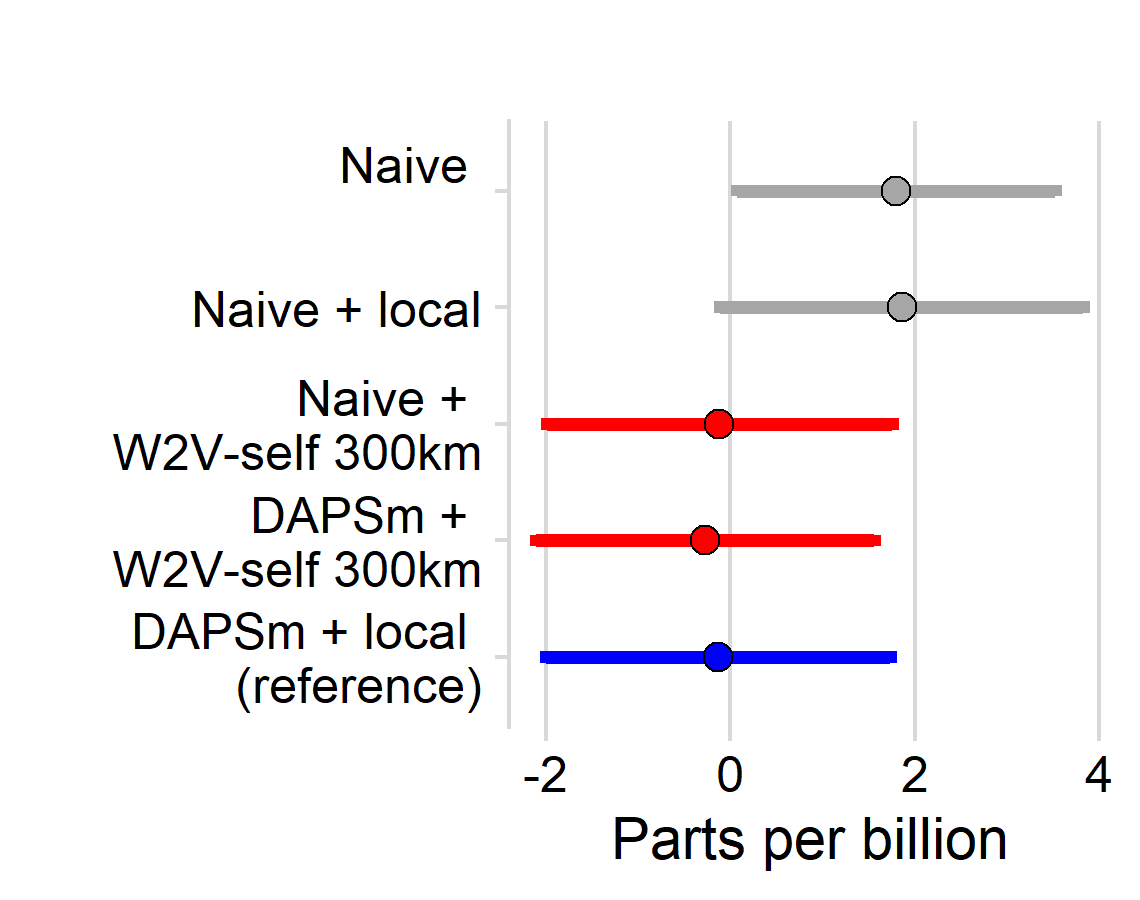}
    \caption{Estimated causal effects}
    \label{fig:app1-effects}
  \end{subfigure} %
  \caption{Application 1: The effectiveness of catalytic devices to reduce power plant ozone emissions.
  }
  \label{fig:app1}
\end{figure*}

\section{Simulation study}\label{sec:simulation}

We conduct a simulation study that roughly mimics a dataset where pollution is dispersed in accordance with non-local meteorological covariates as in our applications. We briefly describe the setup and results here. Appendix \ref{appendix:simulation} contains additional details and visualizations.

\textbf{Data generation summary}. Two-dimensional covariates simulating wind vectors are generated from the gradient field of a random spatial process. The treatment probability and the outcome (simulating air pollution) are non-local functions of the covariates such that areas with lower outcomes have a higher probability of treatment, with a fixed treatment effect of $\tau_\text{ate}=0.1$. Two varying factors are considered: whether $\sS$ is dense or sparse; and whether the simulated data is linear or non-linear on the covariates. The implicit radius of NLC is determined by using $13 \times 13$ convolution kernels to simulate the treatment and outcomes with non-linear operations.

\textbf{Baselines}. We implement the supervised (\textsc{w2v-sup}) and self-supervised methods (\textsc{w2v-self}) using depth-2 U-nets. We then compare them with several baselines: first, no adjustment (\textsc{un}), computed as the difference in means of treated and non-treated; \textsc{local}, which uses local covariates only; \textsc{avg}, which appends averages of neighboring covariates, assuming the neighborhood size is known. Next, we use two convolutional autoencoders baselines of dimension reduction that operate on pre-extracted patches of the oracle size; \textsc{crae} \citep{blier2020encoding} and \textsc{vae} \citep{kingma2013auto}. Notice that although we include these baselines for reference, patch-based estimates do not scale to large datasets. Next, we consider \textsc{wx} linear logistic classification \citep{elhorst2010applied} using a larger kernel of the oracle size. Finally, an approach bosed solely on spatial modeling \textsc{car} \citep{besag1974spatial}, and a hybrid method combining the spatial term with the supervised U-net (\textsc{w2v-car}). 10 random seeds are run for each configuration.

\textbf{Causal estimation.} All the estimates are based on IPTW (Appendix \ref{app:iptw}) from a learned propensity score. For \textsc{local}, {avg}, and methods based on dimension reduction, we fit the learned vectors through a two-layer feed-forward network (FFN) for the propensity score.  We consider four latent dimensions for the self-supervised method and all dimension reduction baselines.

\textbf{Results summary}.  The results are summarized in \cref{tab:simulation}. 
 When $\sS$ is dense, the supervised \method outperforms all others, exhibiting near-zero bias in the linear case and a small amount of finite-sample bias in the non-linear case. The self-supervised version is competitive in all scenarios, performing better than the alternatives in the non-linear sparse case.

\section{Applications in Air Pollution and Climate}\label{sec:applications}

\paragraph{Application 1: Quantifying the impact of power plant emission reduction technologies}

The study aims to quantify the impact of SCR/SNCR catalytic devices \citep{muzio2002overview} to reduce emissions among coal-fired power plants in the U.S \citep{georgia2016dataverse}. Appendix \ref{appendix:app1} provides a description of the dataset. Since air quality regulations are inherently regional and power plants are concentrated in regions with similar weather and economic demand factors, regional weather correlates with the assignment of the intervention. Further, weather patterns (such as wind vectors, precipitation and humidity) dictate regional differences in the formation and dispersion of ambient air pollution. Thus, the weather is a potential confounding factor which cannot be entirely characterized by local measurements.

\begin{figure*}[htb]
  \centering
  \begin{subfigure}[t]{.3\textwidth}
    \centering
    \includegraphics[width=0.99\textwidth]{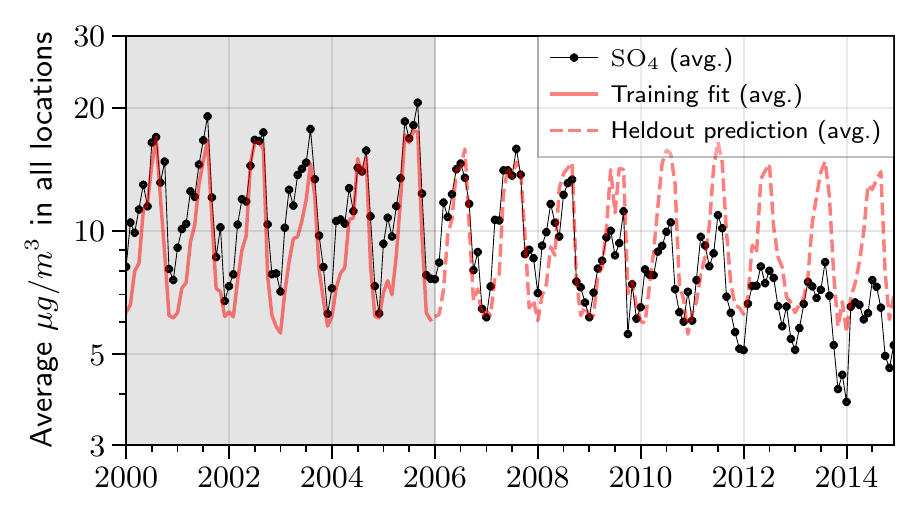}
   \caption{Prognostic score fit averaged over the entire grid $\sG$.}
   \label{fig:app2-prognostic} 
  \end{subfigure}\hfill%
  \begin{subfigure}[t]{.68\textwidth}
    \centering
    \includegraphics[width=0.5\textwidth]{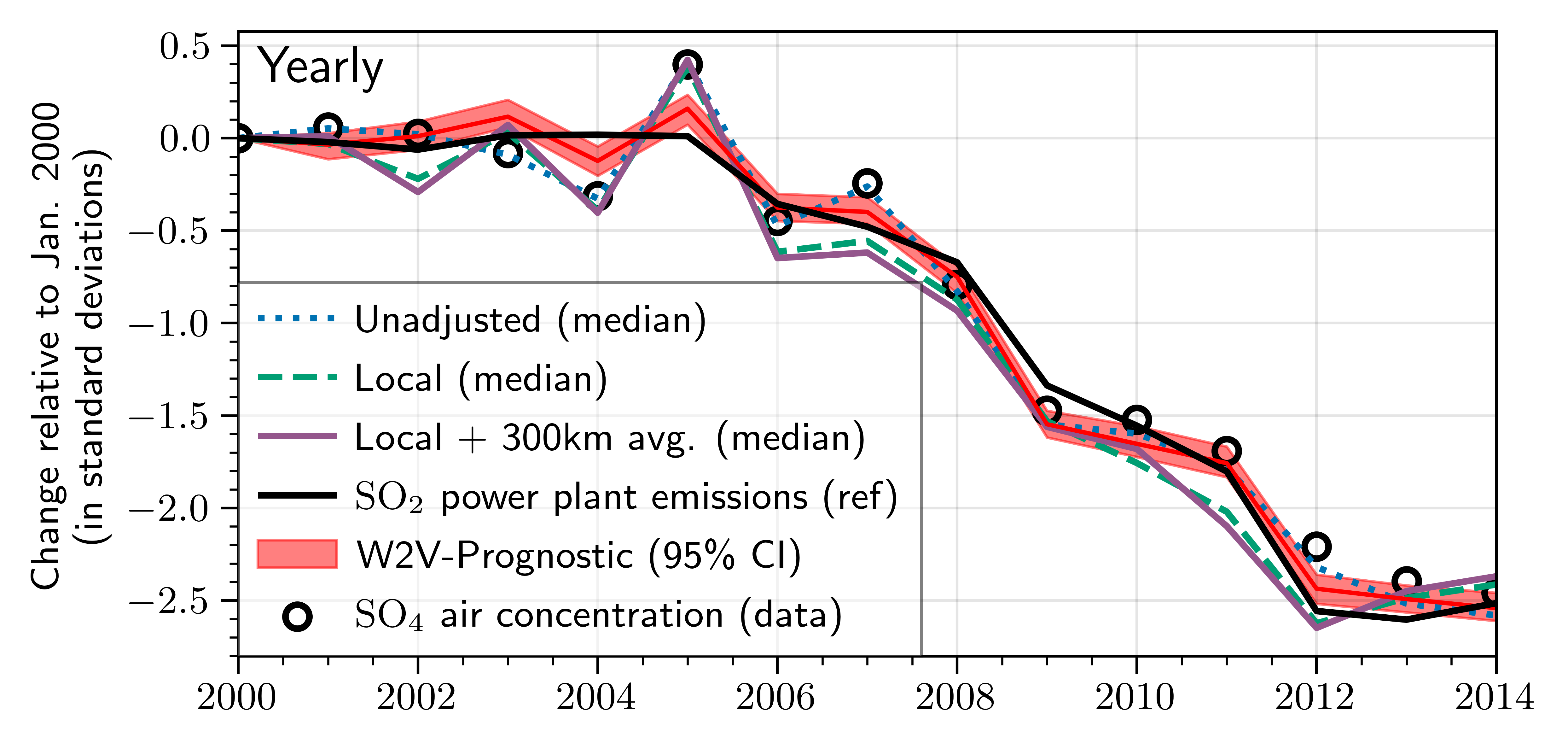}\hfill
    \includegraphics[width=0.5\textwidth]{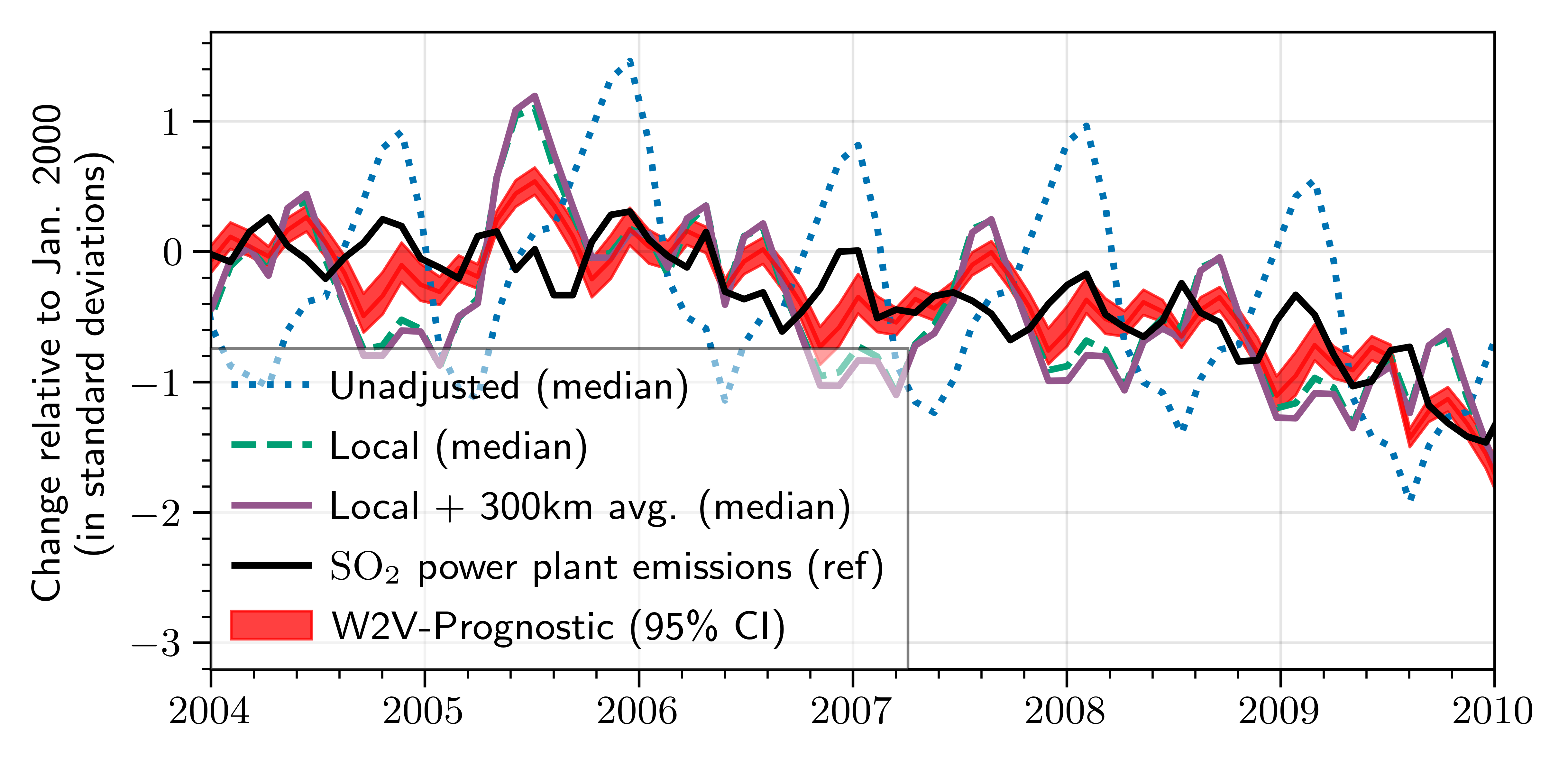}
    \caption{Detrended series at $\sS^*$ resembles power plant emissions. (\emph{Left}) Yearly trend $\delta_{\textrm{year}(t)}$. (\emph{Right}) Monthly trend $\delta_{\text{year}(t)} + \gamma_{\text{month}(t)}$}
    \label{fig:app2-trends}
  \end{subfigure} %
  \caption{Application 2: Meteorological detrending of $\SO_4$.
  }
  \label{fig:app2}
\end{figure*}

\emph{Self-supervised features from NARR.} We construct a dataset of atmospheric covariates following \citet{shen2017influence}. We downloaded monthly NARR data \citep{mesinger2006north} containing averages of gridded atmospheric covariates across the mainland U.S. for the period 2000-2014. We considered 5 covariates: temperature at 2m, relative humidity, total precipitation, and north-sound and east-west wind vector components. For each variable, we also include its year-to-year average. Our dataset is identical to \citet{shen2017influence}, except that they project it to a lower resolution, while we keep it so that each grid cell covers roughly a $32\times 32$ km area, forming a $128 \times 256$ grid. We implemented the self-supervised \method with a lightweight U-net of depth 2, 32 hidden units, and only one convolution per level. See Appendix \ref{appendix:app1} for more details and a schematic of the U-net architecture. To measure the quality of the encoding, \cref{fig:app1-self-supervision} shows the percentage of variance explained ($R^2$), comparing with neighbor averaging and local values. This metric is computed as the coefficient determination, which is essentially the average squared correlation between the prediction and the actual data, aggregated by distance to the center. The results show that the 32-dimensional self-supervised features provide a better reconstruction than averaging and using the local values. For instance, the 300km averages only capture 82\% of the variance, while the self-supervised \method features capture 95\%. See \cref{appendix:app2} for details on the calculation of the $R^2$ and neural network architecture.

\emph{Estimated pollution reduction.}  We evaluate different propensity score models for different neighborhood sizes of the June 2004 NARR \method-learned features with the same logistic model and other covariates as in DAPSm, augmented with the self-supervised features. We selected the representation using features within a 300km radius on the basis of its accuracy, recall, and AIC in the propensity score model relative to other considered neighborhood sizes (Figure \ref{fig:app1-fit-metrics}).  
The causal effects are then obtained by performing 1:1 nearest neighbor matching on the estimated propensity score as in DAPSm. 
Figure \ref{fig:app1-effects} compares treatment effect estimates for different estimation procedures. Overall, standard (naive) matching using the self-supervised features is comparable to DAPSm, but without requiring the additional spatial adjustments introduced by DAPSm. The same conclusion does not hold when using local weather only
, which (as in the most naive adjustment) provides the scientifically un-credible result that emissions reduction systems significantly {\it increase} ozone pollution. Do notice the wide confidence intervals which are constructed using conditional linear models fit to the matched data sets \citep{ho2007matching}. Thus, while the mean estimate shows a clear improvement, the intervals shows substantial overlap, warranting  caution.

\paragraph{Application 2: Meteorological detrending of sulfate}\label{sec:detrending}

We investigate meteorological detrending of the U.S. sulfate ($\SO_4$) time series with the goal (common to the regulatory policy and atmospheric science literature) of adjusting long-term pollution trends by factoring out meteorologically-induced changes and isolating impacts of emission reduction policies \citep{wells2021improved}. We focus on $\SO_4$ because it is known that its predominant source in the U.S. is $\SO_2$ emissions from coal-fired power plants, on which observed data are available for comparison. Thus, we hypothesize that an effectively detrended $\SO_4$ time series will closely resemble that of the power plant emissions. 

\emph{Prognostic score.} We obtained gridded $\SO_4$ concentration data publicly available from \citet{van2021monthly}, consisting of average monthly values in the mainland U.S. in 2000--2014. The data is aggregated into 32km-by-32km cells to match the resolution of atmospheric covariates. The model uses a U-net with quadratic loss for the (log) concentrations of $\text{SO}_4$.
Since the prognostic score is defined based on outcome data in the absence of treatment, we leverage the fact that the power plant emissions were relatively constant for the period 2000-2005 and using 2006 as test data -- regarding this period as absent of treatment. The model predictions, aggregated by all points in the grid is shown in Figure \ref{fig:app2-prognostic}.
The difference between the red line (the prognostic score fit) and the black dotted line (the $\SO_4$) observations during 2000 - 2006 is a proxy for the meteorology-induced changes in the absence of treatment.

\emph{Trend estimation.} For comparability we adhere to the recommended detrending model by \citep{wells2021improved}.  Accordingly, we specify a regression with a year and seasonal fixed-effect term. Rather than pursue an entirely new methodology for detrending, we intentionally adhere to standard best practices and merely aim to evaluate whether augmenting this approach with the \method representation of the prognostic score offers improvement. The outcome $\log(Y_{s,t})$ for untreated units is regressed using the predictive model
\begin{equation}\label{eq:detrending}
\begin{aligned}
\mu_{s,t} &=\alpha + \delta_{\text{year}(t)} + \gamma_{\text{month}(t)} + \textstyle{\sum_{j=1}^p} \beta_{p} X_{st}^p\\
 \end{aligned}
\end{equation}
for all $s\in\sS^*$ and $t=1,\hdots, T$; and where $\delta_\ell$ is the year effect for $\ell=2000,\hdots,2014$; $\gamma_\kappa$ is the seasonal (monthly) effect for $\kappa=1,\hdots,12$; $\sS^* \subset \sS$ are the locations of the power plants; and $X_{st}^p$ are the controls with linear coefficients $\beta_{s,p}$. These controls are obtained from a B-spline basis of degree 3 using: 1) local weather only, and 2) local weather plus the \method prognostic score. The model is fitted using Bayesian inference with a Gibbs sampler. Figure \ref{fig:app2-trends} shows the fitted (posterior median) yearly and monthly trends, which resemble the power plant emissions trends much more closely than the predicted trends from models that include local or neighborhood average weather. Notice the ``double peak'' per year in the monthly power plant emissions (owing to seasonal power demand), which is only captured by the detrended \method series.

\section{Discussion and Future Work}

While notions of NLC have been acknowledged in causal inference (most explicitly in spatial econometrics but also alluded to in literature on spatial confounding and interference), potential-outcomes formalization of NLC and flexible tools to address it are lacking. We offer such a formalization, along with a flexible representation learning approach to account for NLC with gridded covariates and treatments and outcomes measured (possibly sparsely) on the same grid.  Our proposal is most closely tailored to problems in air pollution and climate science, where key relationships may be confounded by meteorological features, and promising results from two case studies evidence the potential of \method to improve causal analyses over those with more typical accounts of local weather. A limitation of the approach is that the learned \method representations are not as interpretable as direct weather covariates and using them could impede transparency when incorporated in policy decisions. Future work could explore new methods for interpretability. Other extensions could include additional data domains, such as graphs and longitudinal data with high temporal resolution. The links to causal interference explored in Section \ref{sec:interference} also offer clear directions for future work to formally account for NLC in the context of estimating causal effects with interference and spill-over.

\section*{Acknowledgments.}
Part of this work was done while Mauricio Tec was a doctoral student at the Department of Statistics and Data Sciences at the University of Texas at Austin. The work was supported by the National Institutes of Health (NIH-R01ES26217, NIH-R01ES030616) and the US Environmental Protection Agency (EPA-RD835872).  Its contents are solely the responsibility of the grantee and do not necessarily represent the official views of the US EPA. Further, the US EPA does not endorse the purchase of any commercial products or services mentioned in the publication. Some of the computations in this paper were run on the FASRC Cannon cluster supported by the FAS Division of Science Research Computing Group at Harvard University.

\bibliography{aaai23}

\clearpage
\appendix
\section*{Appendix}

\section{Inverse Probability of Treatment Weighting (IPTW)}\label{app:iptw}

IPTW is one of the most standard causal inference techniques \citep{rubin2005causal, cole2008constructing}. This estimator emulates a pseudo-population in which confounders are equally distributed between the treated and untreated units. Instead of actually constructing this pseudo-population, it estimates the ATE using weights that are inversely proportional to some propensity score estimate $\hat{\mu}_s=\hat{p}(A_s=1 \mid \mL_s)$ using the formula

\begin{align*}
\hat{\tau}_\mathrm{IPTW}=&|\sS|^{-1}\textstyle{\sum_{s\in\sS}}\{(Y_s/\hat{\mu}_s)\mathbb{I}(A_s=1)\\
& \hspace{6em} - (Y_s/(1 - \hat{\mu}_s))\mathbb{I}(A_s=0) \}.
\end{align*}
When $\gL_s$ is sufficient and the propensity scores are known (instead of estimated), this formula yields unbiased causal estimates of the ATE.

\section{Proofs}\label{appendix:proofs}

\begin{proof}[\bf Proof of Proposition 1]
For convenience, drop the subscript $s$ and boldface notations, and denote $L^c=L' \setminus L$. 
We will use a graphical argument based on the backdoor criterion \citeA[ch. 4.3]{pearl1988probabilistic}. 
Suppose that $L^c \to A$ (here $\to$ means causation) and observe the two following facts: first, a path $Y\to L^c \to A$ would violate the assumption of pretreatment covariates; second, a path $Y \leftarrow L^c \to A$ would need to be absent or be blocked by $L$ due to sufficiency. If blocked, it must be of the form $Y \leftarrow L \leftarrow L^c \leftarrow A$ since a reversed first arrow would violate the pre-treatment assumption, implying $L^c \indep Y \mid L$ (and as a consequence, conditionally independent of  $Y(0),Y(1)$).
An analogous argument shows that assuming $L^c \to Y$ would imply $L^c$ is conditionally independent from $A$ given $L$. In summary, conditioning on $L^c$ does not open any new (backdoor) paths from $A$ to $Y$. And the result follows from the backdoor criterion.
\end{proof}

\begin{proof}[\bf Proof of Proposition 2]
This is a standard result in introductory expositions of potential outcomes. For each $a\in\{0,1\}$ we have that
\begin{align*}
    \E[\E[\out \mid \propconf, \treat=a]] &=
    \E[\E[\out(a) \mid \propconf, \treat=a]]  \\ &=
    \E[\E[\out(a) \mid \propconf]] \\ &= 
    \E[\out(a)].
\end{align*}
The first equality follows from SUTVA; the second from sufficiency; the third from the law of iterated expectation. Finally, $\E(\out(a))=\lvert \sS \rvert^{-1}\sum_s \out(a)$ by definition, implying the proposition's statement.
\end{proof}

\begin{proof}[\bf Proof of Proposition 3] We follow \citeA{hansen2008prognosticA}'s formulation of the prognostic score, and prove the results along the lines of \citeA[theorems 1-3]{rosenbaum1983central}. We'll proceed in three steps. All which are somewhat informative of the role of the prognostic score. Again, we drop the subscript $s$ and boldface from the notation for clarity.

\emph{Step 1. Conditional expectation of the outcome is a prognostic score}. Denote $\psi(L)=\E[Y(0) \mid L]$. We want to show the balancing property: $Y(0) \indep L \mid \psi(L)$.

Recall the definition of conditional expectation (see \citeA[ch. 9.2]{williams1991probability}): $Z=\E[Y\mid L]$ iff $\E[Y\Ind(L\in A)]=\E[Z\Ind(L\in D)]$ for any $L$-measurable set $D$. We will use this definition and show that $p(Y(0) \in C \mid L) = p(Y(0) \in C \mid \psi(L))$, implying the required independence. (Conditioning on $(L,\psi(L))$ is equivalent to only condition on $L$.)

Now, since $\psi(L)$ is a function of $L$, the event $\psi(L)\in D$ can be re-written as $L\in \psi^{-1}(D)$ using the pre-image notation. Then, 
\begin{align*}
& \hspace{-2em} \E[\Ind(Y(0)\in C) \Ind(\psi(L)\in D)] \\ &= \E[\Ind(Y(0)\in C) \Ind(L \in \psi^{-1}(D))]  \\ &= \E[\E[(Y(0)\in C) \mid L]  \Ind(L\in \psi^{-1}(D))],
\end{align*}
implying that $\E[\Ind(Y(0)\in C) \mid L]=\E[\Ind(Y(0)\in C) \mid \psi(L)]$. The result then follows from noting that probabilities are expectations of indicator functions.

\emph{Step 2. Any other prognostic score $b(L)$ is finer than $\psi(L)$.} Suppose it is not the case, then there are $\ell_1, \ell_2$ such that $\psi(\ell_1) \neq \psi(\ell_2)$ but $b(\ell_1)=b(\ell_2)$. But by the balancing property we have that
\begin{align*}
&\hspace{-2em} \E[Y(0) \mid b(L) =b(\ell_1)] \\
&= \E[Y(0) \mid b(L)=b(\ell_1), L=\ell_1] \\
&= \psi(\ell_1),
\end{align*}
which would imply that $\E[Y(0) \mid b(L)=b(\ell_1)]\neq \E[Y(0) \mid b(L)=b(\ell_2)]$, violating the assumption that $b(\ell_1) = b(\ell_2)$ and leading to a contradiction. Thus $\psi(\ell_1)=\psi(\ell_2)$ implies that $b(\ell_1)=b(\ell_2)$, which in turn implies the existence of some function $\psi(L)=f(b(L))$ and thus $\psi(L)$ is coarser.

\emph{Step 3. If $b$ is a prognostic score, then $L$ is sufficient iff $b(L)$ is also sufficient}. First, if $b(L)$ is sufficient, then the proof is trivial. So let's consider the opposite case.  First we show that $p(Y(0)\in C \mid A, b(L))=p(Y(0) \in C \mid b(L))$. 

The proof follows from the following identities
\begin{align*}
&\hspace{-2em}p(Y(0)\in C \mid A, b(L))\\
&=\E[\Ind(Y(0)\in C) \mid b(L)] \\ &= \E[\E[\Ind(Y(0)\in C) \mid L]  \mid A, b(L)] \\ &= \E[\E[\Ind(Y(0)\in C) \mid \psi(L)] \mid A, b(L)] \\
& = \E[\Ind(Y(0)\in C) \mid \psi(L)] \\
&= p(Y(0)\in C \mid b(L)).
\end{align*}
The first equality is by definition, the second by iterated expectation, the third one by the sufficiency of $L$; the fourth one is because $\psi(L)$ is balancing (Step 1); the fifth one is because $\psi(L)$ is a function of $b(L)$ by Step 2; the last one is by definition.

Finally, for the treated outcome $Y(1)$, the assumption of no effect modification means that the same argument carries on for $Y(1)$ (since $Y(1) - Y(0)$ is independent of $A$). 
\end{proof}

\begin{proof}[\bf Proof of Proposition 4]
Let $a\in\{0,1\}$. By the assumption of conditional independence of the treatments given $\propconf$ (assumption (2) in the proposition), we have that
\begin{align*}
    \E[\out \mid \propconf, \treat=a]] &= \E[\out \mid \propconf, \treat=a, \tA_{\neighminus}]
\end{align*}
Having noted this, the proof is identical to that of Proposition 2
\begin{align*}
   & \hspace{-0.5em} \E[\E[\out \mid \propconf, \treat=a, \tA_{\neighminus}]] \\ 
   &= \E[\E[\out(a_s=a, \tA_{\neighminus}) \mid \propconf, \treat=a, \tA_{\neighminus}]]  \\
   & =
   \E[\E[\out(a_s=a, \tA_{\neighminus}) \mid \propconf]] \\
   &= \E[\out(a_s=a, \tA_{\neighminus})]
\end{align*}
where the first identity is due to SUTNVA; the second one is by neighborhood-level sufficiency (assumption (1) in the proposition); and the third one is by the law of iterated expectation. Finally, $E[\out(a_s=a, \tA_{\neighminus})]=(1/|\sS|)\sum_s \out(a_s=a, \tA_{\neighminus})$ since the randomness in the expectation is due to $s$ uniformly from $\sS$.
\end{proof}

\section{Motivating example for the self-supervised model: perfect encoding}
\label{appendix:motivation-selfsup}

Assume that the covariates $\covar$ have dimension $d=1$ and that the self-supervision task is to learn the adjacent values in the grid (north, west, south, east) and the central point $s=(i,j)$ using the representation $\lat$. If we set the representation dimension to $k=5$, then the obvious candidate for the representation is
$$
\mZ_{(i,j)} = (\mX_{(i-1,j)}, \mX_{(i,j-1)}, \mX_{(i+1,j)}, \mX_{(i,j+1)}, \mX_{(i,j)})^\top
$$
Now let $\vgamma(\ell)=(\vgamma(\ell)_1,\hdots,\vgamma(\ell)_5)$ be the $\ell$-th indicator vector with $\vgamma(\ell)_j=\Ind(j=\ell)$. Then
\begin{align*}
 \vgamma(1)^\top  \mZ_{(i,j)}= \mX_{(i-1,j)},  \cdots \quad\quad \vgamma(5)^\top \mZ_{(i,j)} = \mX_{(i,j)}
\end{align*}
Hence $\mZ$ is a perfect encoding and the $\vgamma(\ell)$'s are perfect classifiers for each offset $\ell$. To generalize this idea to higher dimensions $d > 1$, we can take $\vgamma(\ell)$ to be a $d\times k$ matrix for each $\ell$. Then $\vgamma(\ell)^\top \lat$ is a $d$-dimensional vector for each offset $\ell$. The same idea is behind the self-supervised model, which takes $\mGamma = \gamma^\top$ as a $k\times k$ matrix and adds a decoder neural network. Rather than using indicator functions for $\Gamma$, the method formulates it as a neural network that is a function of the offset.

\section{A connection between the self-supervised model and PCA}\label{appendix:pca}

Principal components analysis (PCA) is closely related to a special case of the self-supervised \method when using a single $(2R+1)\times(2R+1)$-convolution instead of the U-net, leaving $h_\psi$ as the identity function, and defining the offset embedding $\mGamma(\delta)$ as independent $d \times k$ vectors for each offset $\delta$ (rather than a neural network with $\delta$ as a continuous input). The equivalence is in the sense of reconstruction since both methods can be seen as minimizing the reconstruction error. However, in the self-supervised case there is no guarantee that the latent dimensions of $\lat$ will be orthogonal as in PCA applied to each patch of size $(2R+1)\times(2R+1)$.

\section{Software and Hardware}\label{sec:computing}
We use open-source software PyTorch 1.10 \citeA{NEURIPS2019_9015} on Python 3.9 \citeA{van1995python} for training all the models on a single laptop with an Nvidia GPU 980M (8GB) and a CPU Intel i7-4720HQ at 2.60GHz. The code uses fairly standard functions for NN training and we did not attempt to optimize it for speed. We also use R 3.6 \citeA{r-base} for downloading and pre-processing atmospheric data from NARR, as well as for comparison with DAPSm in application 1 (see Appendix \ref{appendix:app1}).

The code for Bayesian inference in Application 2 is implemented in pure Python as a straightforward Gibbs sampler since the model is Gaussian.

\section{Details of the simulation study}
\label{appendix:simulation}

\begin{figure*}[tbh]
  \centering
  \includegraphics[width=0.95\textwidth]{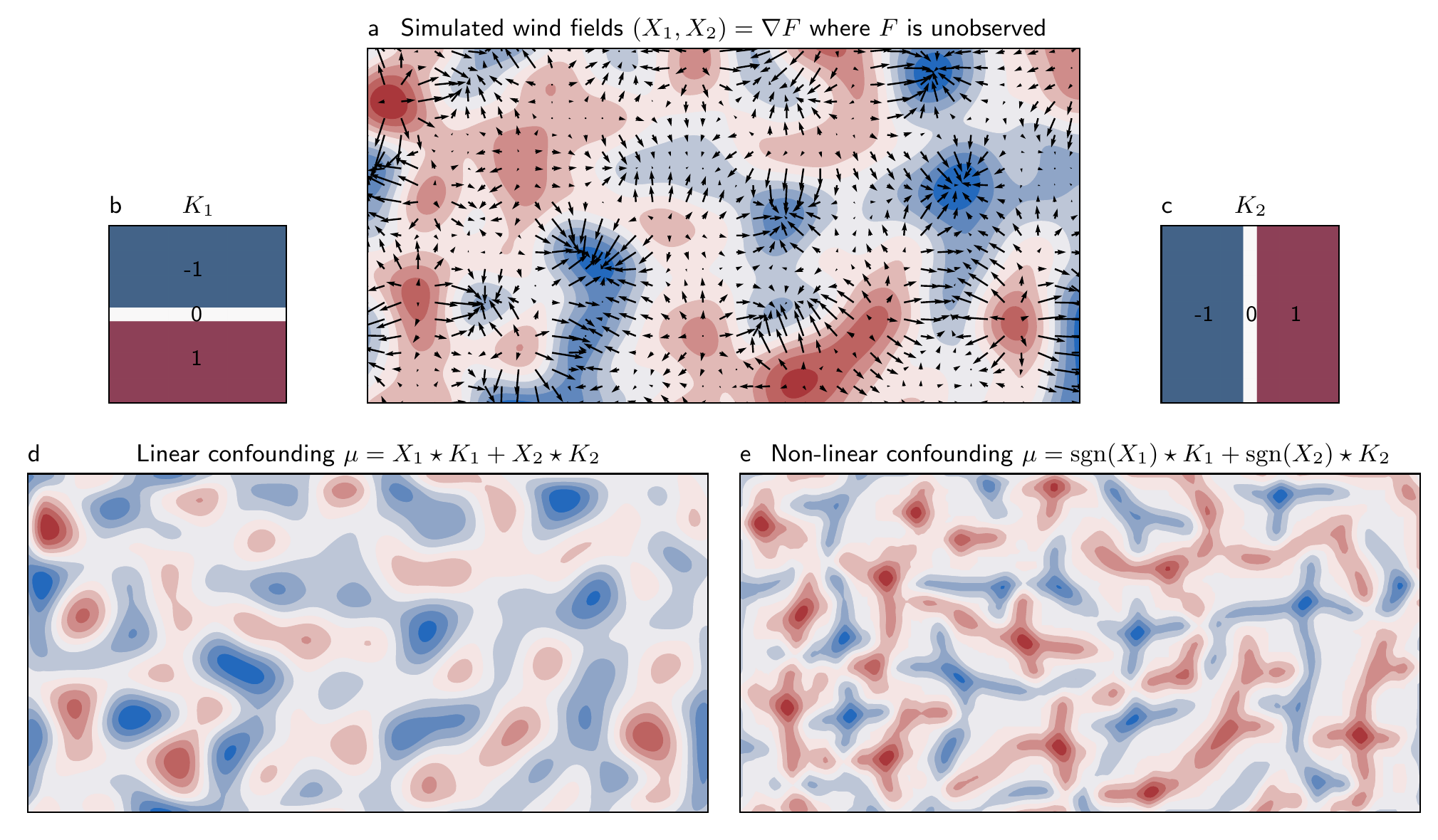}
  \caption{Simulations, components and variants in the simulation study}
  \label{fig:potential}
\end{figure*}

The simulated data mimics the meteorological data in our applications and the matches setup of Section \ref{sec:rubin} with SUTVA and NLC. 

\emph{Data simulation and basic linear task.} The covariates $\covar$ are the gradient field (the first differences along rows and columns) of an unobserved Gaussian Process \citep{rasmussen2003gaussian} defined over a $128\times 256$ grid.  To fix ideas, the simulation is carried out to roughly mimic a study of pollution sources where pollution is dispersed in accordance with  non-local weather covariates, so, $\covar=(\mX_{s}^1, \mX_{s}^2)$ can be roughly interpreted as ``wind vectors". The treatment assignment probability (the propensity score) and the outcome are computed as a ``non-local'' function of $\covar$, simulated to correspond to higher probability of treatment in areas that tend to disperse more pollution. Such an assignment can be performed using a convolution operation. More precisely, let $\mu=\sum_{j\in\{1,2\}} K_j \star \tX^j$ be the result of convolving $\tX$ with a specially designed convolution kernel $K$. Then, the treatment assignment probability is $\treat \sim \mathrm{Bernoulli}(\mu_s)$ and the outcome is $\out = -\mu_s + \epsilon_s + \tau \treat$, where $\tau$ is the treatment effect and $\epsilon_s$ is a mixture of spatial and random noise of unit variance. $\tau=0.1$ in all experiments. $K$ has dimensions $13 \times 13$ (its size determines the radius of NLC). $K_1$ contains -1's in the upper half, +1's in the lower half, and 0's in the middle row. $K_2=K_1^\top$. Convolving a a gradient field with $K$ is an approximate form of identifying valleys and hills in the potential of the gradient field. In this basic formulation, $\sS=\sG$, meaning that $\treat$ and $\out$ are densely available over the grid $\sG$.

\emph{Additional task variants}. 
In one variant, we consider a sparse configuration in which the outcome and treatment are only sparsely available in a subset $\sS$ of 500 randomly selected points in $\sG$. In another variant, we evaluate the results on a \emph{non-linear} version of the treatment assignment logits, computed as $\mu=\sum_{j\in\{1,2\}} K_j \star \mathrm{sign}(\tX^j)$. This small amount of non-linearity strongly increases the complexity of the problem. Both tasks variants are also combined, resulting in 4 total tasks.

Figure \ref{fig:potential} illustrates the data used in the simulation study. (a) shows the simulated covariates $\tX=(\tX^1, \tX^2)$ as the gradient vector field of an unobserved potential function (sampled from a Gaussian Process) $\tF$, whose level curves overlay the covariates (vector field), represented by arrows. (b) and (c) jointly compose the kernel used to generate the confounding factor. (d) is the resulting treatment assignment probability for the linear task, and (e) is the corresponding non-linear variant. In both, convolutions approximately correspond to valleys and hills of the unobserved potential.

\emph{Causal estimation procedure.} 
We first estimate a propensity score model $\hat{\mu}_s$ using the learned $\lat$. For the supervised variant, $\hat{\mu}_s=\mathrm{sigmoid}(\lat)$. But for the self-supervised, it is constructed from an feed-forward network with one or two hidden layers. In all cases, the final estimate of $\tau$ is produced using IPTW.
Although more sophisticated causal estimators could be used, the exercise only intends to measure the degree to which a propensity score anchored to $\lat$ encodes the necessary NLC information.

\begin{figure*}[tbh]
  \centering
  \includegraphics[width=0.8\textwidth]{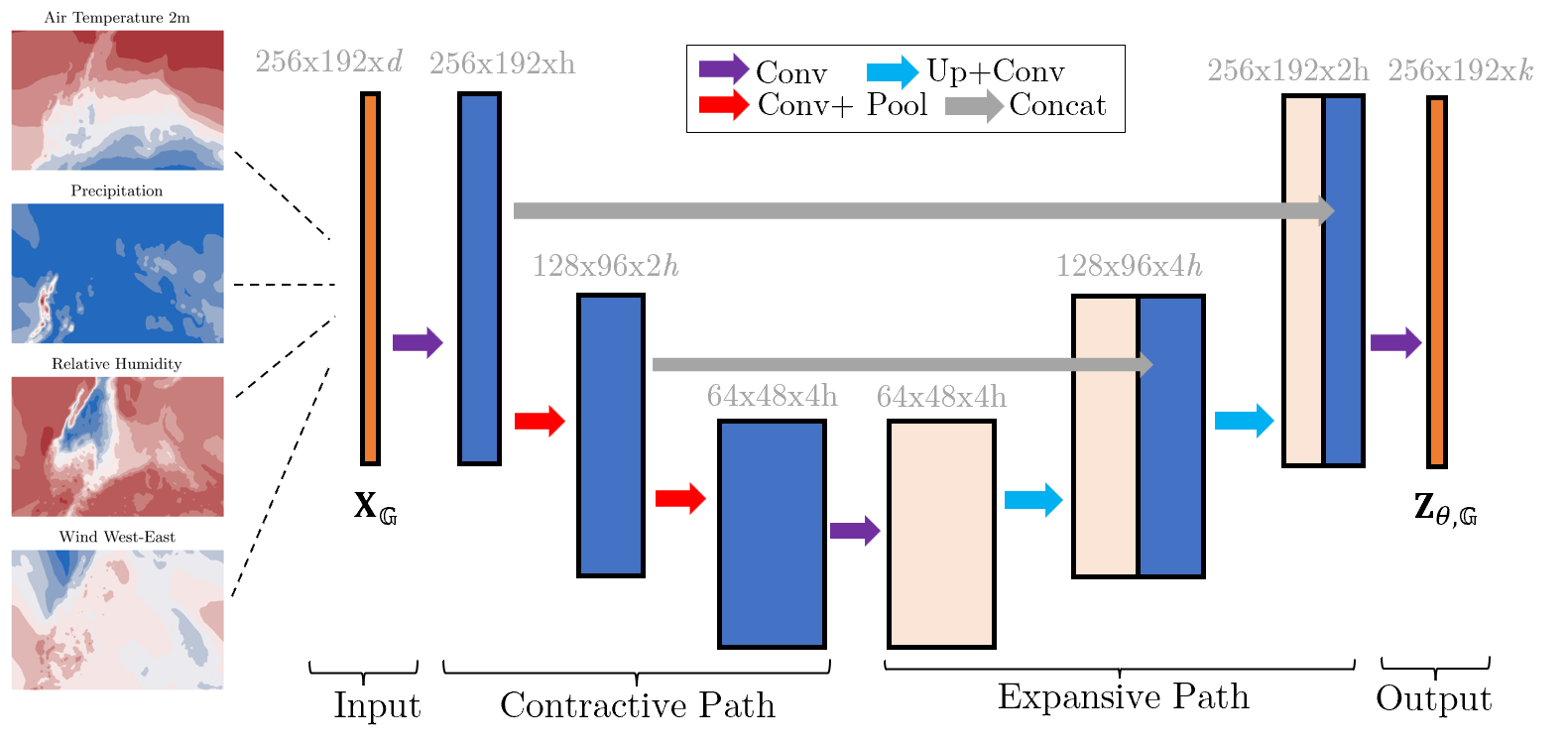}
  \caption{Basic U-net architecture used in the two applications and the simulation study.}
  \label{fig:architecture}
\end{figure*}

\emph{Details on the neural network architectures.} The NNs used in the study are very lightweight, since the data consists of only one image of $128x256$. Typical NN sizes with millions of parameters would easily overfit to this task. The basic U-net architecture used for \method is in Figure \ref{fig:architecture}, but using the simulated gradient fields instead of atmospheric covariates. All convolutions and linear layers are followed by batch normalization and SiLU activations \citeA{elfwing2018sigmoid}, except in the last layer.

\begin{itemize}
    \item {\bf Supervised \method}. The propensity score model uses two hidden units and depth 2. The model has 1.2k (trainable) parameters. 
    \item {\bf Self-supervised \method}. The auto-encoder uses 16 hidden units and depth two. The offset model $\mGamma_\phi$ is a two-layer feed-forward network with 16 hidden units. The decoder $h_\psi$ is feed-forward network with one hidden layer of also 16 units. In total, the auto-encoder has 77k parameters. In addition, the propensity score model uses a feed-forward network with two hidden layers of 16 units, resulting in 600 parameters.
    \item {\bf Local and Local$+$Averages.} These baselines use the same propensity score model as the self-supervised one. Due to their smaller input size, they have around 400 parameters.
    \item {\bf Spatial RE}. Rather than a neural network, we used a conditionally auto-regressive (CAR) \citeA{besag1974spatialA} model such that $A_s \sim \mathrm{Bernoulli}(\mathrm{sigmoid}(\lat))$, $\lat \sim \mathrm{CAR}(\lambda)$ and $\lambda \sim \mathrm{Gamma}(1,1)$. The CAR portion of the negative loglikelihood penalizes the (squared) differences of adjacent values of $\lat$ in the grid by a factor of $\lambda$. Notice that $\lambda$ here is learned along with the model. We remark that CAR models are more scalable alternatives to Gaussian process for applications requiring only smoothing and interpolation. 
    \item {\bf Supervised \method + spatial RE}. This variant formulates the representation as $\lat = \tilde{Z}_{\theta,s} + \xi_s$, where $\tilde{\mZ}_{\theta,s}$ is the output of the U-net and $\xi_s$ has a CAR prior and is restricted to $\sum_s \xi_s = 0$ for identifiability. Intuitively, the term $\xi_s$ captures the errors in the propensity score model that have a strong spatial distribution.  
\end{itemize}

\emph{Details on the training procedures and hyper-parameters.} 
In all cases, we use a fixed learning rate of $10^{-4}$, a weight decay of $10^{-4}$, and 20,000 gradient steps with the ADAM optimizer \citep{kingma2014adam}. The full simulation study takes about 8 hours to finish running two baselines in parallel. The values of weight decay, training epochs and learning rate were chosen as reasonable values without much additional optimization. The number of layers and architectures were chosen by inspection after a few runs, aiming to find a model small enough as to avoid over-fitting without requiring tuning the regularization hyper-parameters or early stopping.

\section{Additional details of Application 1}
\label{appendix:app1}

\emph{Atmospheric data download}. The NARR \citepA{mesinger2006northA} dataset associated with the application can be downloaded via FTP with the R script provided in the code accompanying the paper. The data is also publicly available for download from the website of the National Oceanic and Atmospheric Administration (NOOA) \citeA{nooa}. We could not find any license information for the dataset. We could not find any license attached to the dataset.

\emph{Power plants data}.  Information for the largest 473 coal-fired power plants emitting $\text{SO}_2$ during 2000--2014 was obtained from \cite{georgia2016dataverse} (publicly available under creative commons license CC0 1.0). 

\emph{Neural network architecture for self-supervised features}.  The auto-encoder uses 32 hidden units and depth 3. The offset model $\mGamma_\phi$ is a two-layer feed-forward network with 32 hidden units. The decoder $h_\psi$ is feed-forward network with one hidden layer of also 32 units. In total, the auto-encoder has 1.2M parameters. The architecture is shown in Figure \ref{fig:architecture}. Convolutions are followed by FRN normalization layers \cite{singh2020filter} and SiLU activations. Pooling uses \emph{MaxPool2d} and upsampling use \emph{Bilinear Upsampling2d} as implemented in PyTorch. The model architecture was not tuned since the model with 32 hidden units seemed to work well.

\emph{Details on the training procedures and hyper-parameters.} 
The model is trained for 300 epochs using batch size 4, a linear decay learning rate from $10^{-2}$ to $10^{-4}$ using the ADAM \citep{kingma2014adamA} optimizer (no weight decay). The number of epochs and learning rate were tuned by inspection after a few runs simply to ensure the model was learning at a reasonable speed, but not tuned otherwise. The atmospheric covariates were standardized before training.

We do not split in training and validating datasets since the model is a compression/dimensionality reduction technique, and thus it cannot over-fit. (In fact, an ``over-fitting" here would be a desirable property, since it would mean a perfect dimensionality reduction.)

\emph{Computation of the explained variance ($R^2$).} The traditional $R^2$ is defined as one minus the ratio of sum-of-squares between the prediction errors and the centered targets. Since the covariates are standardized, the latter quantity is simply $N$.
In each training epoch we collect the sum of squared prediction errors for all time periods. Denote this quantity as $SSE_j$ where $j$ indicates the covariate dimension for $j=1,...,d$. Then $R_2= 1 - (Nd)^{-1}\sum_j SSE_j$ is the proposed estimator of the fraction of the variance explained.

\emph{Comparison with DAPSm}. We modified the DAPSm authors implementation from Github \citep{dapsmA} (no license provided) to include the \method self-supervised features as another predictor in their otherwise unchanged propensity score model. The modified R script is in the code accompanying this paper. 

\section{Additional details of Application 2}
\label{appendix:app2}

\emph{Neural network architecture for supervised features}. The U-net architecture for the prognostic score model follows the same general architecture as application 1 (figure \ref{fig:architecture}), but with dimension one in the last layer ($k$). In addition we use depth 2 and $d=8$ (hidden features multiple) to reduce the number of parameters since we empirically observed overfitting with the same parameters as the first application. However, we do not conduct an explicit hyper-parameter sweep. We used the period 2000-2005 as training data (because emissions are mostly flat, see figure \ref{fig:app2-trends}) and choose the model weights that minimize the test mean squared error using data from 2006.

\emph{$\SO_4$ data download}. We downloaded the dataset the $\SO_4$ grid for inland US from the website of the Atmospheric Composition Analysis Group's \citeA{van2021monthlyA} website  \citeA{so4dataA}. We could not find any license information for the dataset. Instructions for replications are provided in the code.

\emph{Missing data}. Data for some observations in the  $\SO_4$ grid are missing and a few have clearly erroneous (near infinite) values. In addition, some locations have data but present zeros through the entire period. We excluded these values using a binary mask in the likelihood by removing non-finite values and keeping only locations with positive observations throughout. Doing so greatly improved the quality of the fitted model. The final locations cover most of the inland U.S., with missing areas mostly outside the U.S, oceans, or the Rocky West and Great Basin. 

\emph{Details on the training procedures and hyper-parameters.} 
The hyper-parameters are also the same as in the self-supervised model except that we use a weight decay of $10^{-4}$ to reduce over-fitting. We did not tune this parameter, however, we did not notice a significant difference by increasing or decreasing its value by a factor of 10.

\section{Visualization of the U-net receptive field}

To visualize the ability of the U-net to capture non-local dependencies we conducted an experiment to visualize the receptive fields of the U-net using data from Application 2 where the outcome $\mY_t$ is $\SO_4$ for the years 2000--2014 across the mainland U.S. and the predictive variables $\tX_t$ are the NARR atmospheric covariates. We fitted a regression model using the U-net for varying depths in $\{1,\hdots, 5\}$. To avoid an explosion of the latent parametters and overfitting, we do not duplicate the number of latent features in the contractive path of the U-net, and instead keep it constant at $32$. Finally, we estimate the receptive field $RF(\cdot)$ using standard gradient techniques \citep{luo2016understanding} by choosing a target output location $s_*$ over the grid, and compute the average gradient (across time points) of the output $Y_{t,s_*}$ with respect to inputs $\mX_{t,s}$ from all locations over the grid as 
$$
RF(s) = \frac{1}{180} \sum_{t=1}^{180}  \left\lVert \frac{\partial Y_{t,s_*}}{\partial\mX_{t,s}} \right\rVert_2.
$$ 
\begin{figure}
    \centering
    \includegraphics[width=\columnwidth]{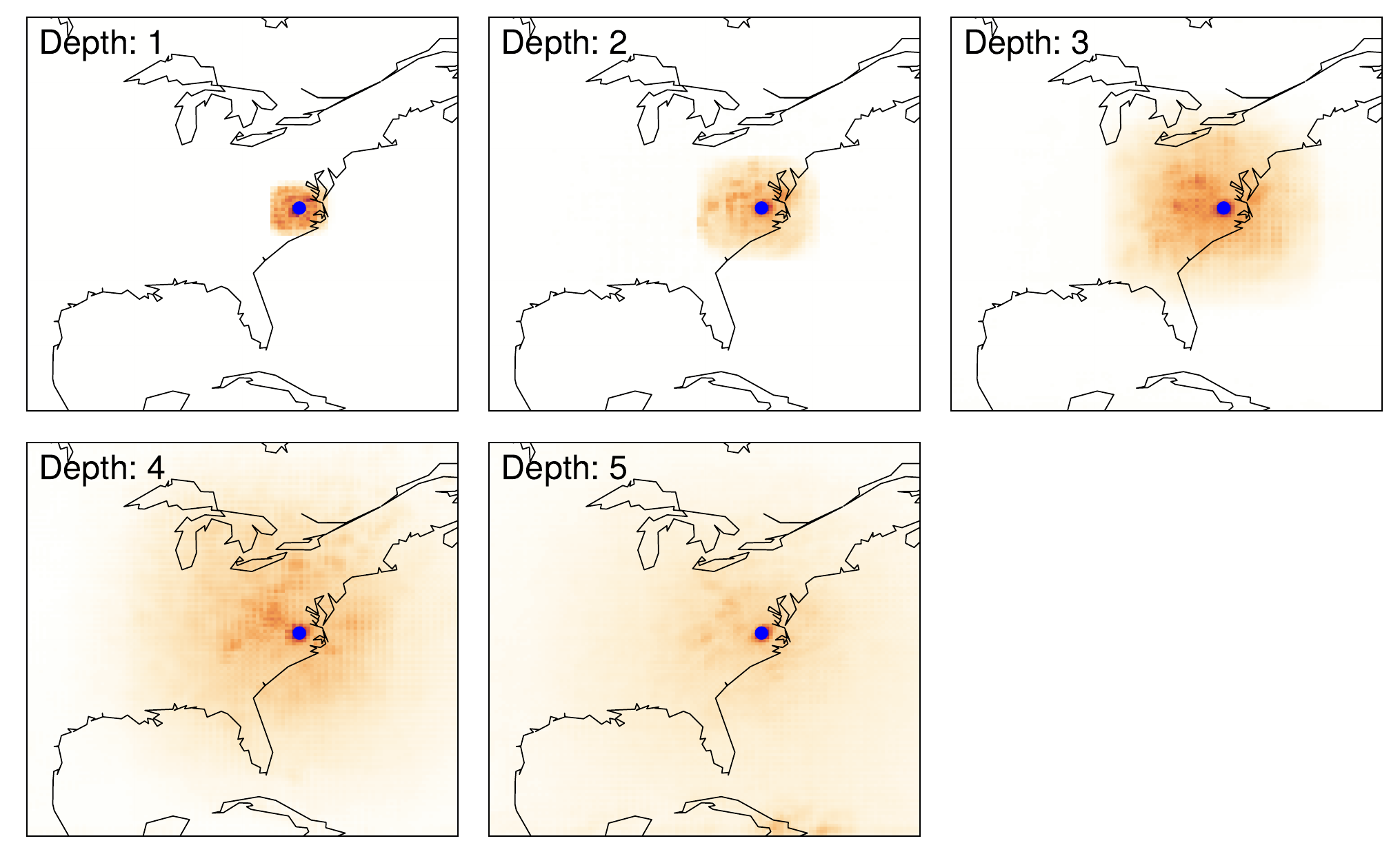}
    \caption{Receptive field by varying depths of the U-net.}
    \label{fig:receptive-fields}
\end{figure}
\cref{fig:receptive-fields} show the results. The orange color indicates a higher gradient. The target outcome location $s^*$ is shown as a blue dot. We can see that as the depth increases, so does the receptive field around the point. However, after depth 3, increasing to depth 4 maintains a higher focus on the same areas west to the target location. This observation suggests that the U-net learned the effective sources of non-local correlation with the outcome. 

\begin{figure}
    \centering
    \includegraphics[width=.7\columnwidth]{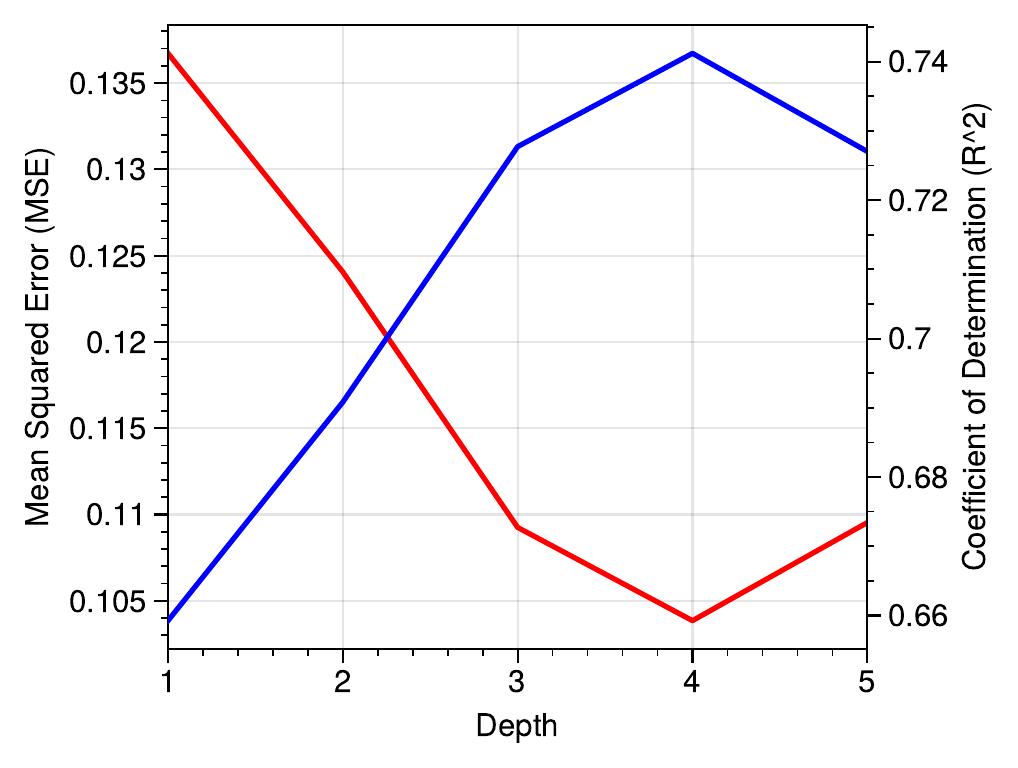}
    \caption{Performance in validation set by varying depths.}
    \label{fig:performance-depths}
\end{figure}

The case when the depth is 5 is also interesting, since it shows a potential problem when setting the depth too high. Some areas of spurious correlation appear (for example in the lower border of the image). This phenomenon could be due to padding effects since, after 5 contractive layers, the dimension $128\times 256$ is reduced to $4\times 8$.  The previous observations are confirmed in \cref{fig:performance-depths} by looking at the out-of-sample performance by masking 10\% of the training data when fitting the model. The performance metrics suggest that in this case, the optimal depth would be 4; although as we previously observed, increasing from depth 3 to depth 4 has a smaller effect.

\section{Algorithms}\label{sec:algorithms}

\cref{alg:sup,alg:self} summarize the two main approaches to learning representation of NLC using the U-net.

\begin{algorithm}[!hbt]

\caption{Supervised representation learning and causal estimation}
\label{alg:sup}
\textbf{Input}: Covariates $\mX_{\sG}$; outcome and treatment $\out$ and $\treat$ at a dense subset $\sS\subset \sG$; and U-net model $f_\theta$ with 1-dimensional outputs. 
\begin{algorithmic}[1] %
\STATE Obtain neural network weights $\hat{\theta}$ minimizing the supervised loss (equations \ref{eq:sup-prop} or \ref{eq:sup-prog}) and compute output balancing score grid from optimal weights $\hat{\tZ}_{\sG}=f_{\hat{\theta}}(\mX_{\sG})$.
\STATE Use standard causal inference methods (e.g., IPTW in Appendix \ref{app:iptw}) to obtain $\hat{\tau}_\text{ATE}$ adjusting for $\hat{\mZ}_{s}$ and other relevant local confounders  at each unit $s$.
\end{algorithmic}
\end{algorithm}

\begin{algorithm}[!bth]
\caption{Self-supervised non-local representation learning for causal estimation}
\label{alg:self}
\textbf{Input}: Covariates $\mX_{\sG}$; outcome and treatment $\out$ and $\treat$ at any subset $\sS\subset \sG$; U-net model $f_\theta$ with $k$-dimensional outputs; and grid of candidate radii $\gR=\{R_1,\hdots,R_\text{max}\}$. 
\begin{algorithmic}[1] %
\STATE For each $R \in \gR$, obtain neural network weights $\hat{\theta}^R$ minimizing the self-supervised loss (equations \ref{eq:self}) and compute $\hat{\tZ}_{R, \sG}=f_{\hat{\theta}^R}(\mX_{\sG})$.
\STATE Use a propensity/prognostic score model to choose the optimal $r$. For example, use a logistic regression taking inputs $\hat{\mZ}_{R,s}$ (and other relevant local confounders) and compute the Akaike information criterion (AIC). Then choose $\hat{r}$ that minimizes the AIC.
\STATE Estimate the $\tau_\text{ATE}$ using the learned propensity score model, e.g., using IPTW (Appendix \ref{app:iptw}).
\end{algorithmic}
\end{algorithm} 

\FloatBarrier

\bibliographystyleA{aaai23}
\bibliographyA{aaai23copy}

\end{document}